\documentclass[letterpaper, 10 pt, journal, twoside]{IEEEtran} % Use this command for final RAL version

\IEEEoverridecommandlockouts                              % This command is only needed if 
\usepackage{amsmath,amsfonts}
\usepackage{algpseudocode}
\usepackage[ruled,vlined,linesnumbered]{algorithm2e}
\usepackage{array}
\usepackage[caption=false]{subfig}
\usepackage{textcomp}
\usepackage{stfloats}
\usepackage{url}
\usepackage{verbatim}
\usepackage{graphicx}
\usepackage{cite}
\usepackage{bm}
\usepackage{color}
\usepackage{xcolor}
\usepackage{multirow}
\usepackage{booktabs} 
\usepackage{tabularx}
\usepackage{lipsum}  
\usepackage[export]{adjustbox}
\usepackage{subfig}

\makeatletter
\let\NAT@parse\undefined
\makeatother
\usepackage{hyperref}
\hypersetup{
    colorlinks=true,
    linkcolor=black,
    filecolor=black,      
    urlcolor=black,
    citecolor=black,
    pdftitle={Real-time Planning of Minimum-time Trajectories for Agile Flight},
    }

\newcommand{\PREPRINTYEAR}{2024}
\newcommand{\PUBLISHEDIN}{IEEE Robotics and Automation Letters}
\newcommand{\DOI}{10.1109/LRA.2024.3471388}

\newcommand{\changed}[1]{{\color{black}#1}}
\newcommand{\change}[1]{{\color{black}#1}}

\newcommand{\reffig}[1]{Fig.~\ref{#1}}

\newcommand{\refalg}[1]{Alg.~\ref{#1}}
\newcommand{\refsec}[1]{Sec.~\ref{#1}}
\newcommand{\reftab}[1]{Table~\ref{#1}}
\newcommand{\refeq}[1]{\eqref{#1}}
\SetKwInput{KwData}{Global}

\title{Real-time Planning of Minimum-time Trajectories for Agile UAV Flight}

\author{Krystof Teissing$^*$, Matej Novosad$^*$, Robert Penicka, Martin Saska% <-this % stops a space
\thanks{Manuscript received: July, 1, 2024; Revised August, 28, 2024; Accepted September, 19, 2024.%Use only for final RAL version
This paper was recommended for publication by Editor G. Loianno upon evaluation of the Associate Editor and Reviewers' comments.}
\thanks{The authors are with the Multi-robot Systems Group, Faculty of Electrical
Engineering, Czech Technical University in Prague, Czech Republic (\protect\url{http://mrs.felk.cvut.cz/}). 
This work has been supported by the Czech Science Foundation (GA\v{C}R) under research project No. 23-06162M, by the European Union under the project Robotics and advanced industrial production (reg. no. CZ.02.01.01/00/22\_008/0004590), and by CTU grant no SGS23/177/OHK3/3T/13.}%% <-this % stops a space
\thanks{Digital Object Identifier (DOI): see top of this page.}
}

\usepackage[placement=top,vshift=-7]{background}
\SetBgScale{1.0}
\SetBgContents{\PUBLISHEDIN. PREPRINT VERSION - DO NOT DISTRIBUTE. \href{https://doi.org/\DOI}{DOI \DOI}}
\SetBgColor{black}
\SetBgAngle{0}
\SetBgOpacity{1.0}

\begin{document}

% Place this in the document body as the first part to have an extra first page with the copyright. Remove this part to get rid of the extra page.
\thispagestyle{empty}
\onecolumn
{
  \topskip0pt
  \vspace*{\fill}
  \centering
  \LARGE{%
    \copyright{} \PREPRINTYEAR~\PUBLISHEDIN\\\vspace{1cm}
    Personal use of this material is permitted.
    Permission from \PUBLISHEDIN~must be obtained for all other uses, in any current or future media, including reprinting or republishing this material for advertising or promotional purposes, creating new collective works, for resale or redistribution to servers or lists, or reuse of any copyrighted component of this work in other works.}
    \vspace*{\fill}
}
\NoBgThispage
\twocolumn          	% Comment out for single-column articles
\BgThispage

\maketitle
\def\thefootnote{*}\footnotetext{These authors contributed equally to this work.}
% \thispagestyle{empty}
% \pagestyle{empty}

%%%%%%%%%%%%%%%%%%%%%%%%%%%%%%%%%%%%%%%%%%%%%%%%%%%%%%%%%%%%%%%%%%%%%%%%%%%%%%%%
\begin{abstract}

\changed{
We address the challenge of real-time planning of minimum-time trajectories over multiple waypoints, onboard multirotor UAVs.
% \robert{what about multiwaypount add above?}
Previous works demonstrated that achieving a truly time-optimal trajectory is computationally too demanding to enable frequent replanning during agile flight, especially on less powerful flight computers. % be executed online.
% Our approach overcomes this stumbling block by utilizing a point-mass model with a novel iterative thrust decomposition algorithm, which maximizes the UAV's collective thrust and surpasses traditional conservative per-axis acceleration constraints. 
Our approach overcomes this stumbling block by utilizing a point-mass model with a novel iterative thrust decomposition algorithm, enabling the UAV to use all of its collective thrust, something previous point-mass approaches could not achieve.
% \robert{the goal is not to utilize the max collective theust but to be able to use it, which is something the other pmms were not able to}
The approach enables gravity and drag modeling integration, significantly reducing tracking errors in high-speed trajectories, which is proven through an ablation study.
% When combined with a new gradient-based multi-waypoint velocity optimization algorithm, the proposed method generates minimum-time multi-waypoint trajectories within milliseconds. 
When combined with a new multi-waypoint optimization algorithm, which uses a gradient-based method to converge to optimal velocities in waypoints, the proposed method generates minimum-time multi-waypoint trajectories within milliseconds. 
% \robert{maybe add some hint how the multiwaypoint works....}
% We demonstrate that trajectories generated by the proposed method yield similar or even smaller tracking errors than the trajectories generated for a full multirotor model, when flown using Nonlinear Model Predictive Control (NMPC) both in simulation and in real-world, with acceleration of up to 3.5g and velocities over 100 km/h. 
The proposed approach, which we provide as open-source package, is validated both in simulation and in real-world, using Nonlinear Model Predictive Control. 
% \matej{would it be over the line to just use NMPC abbreviation over full name?}
% \robert{i would remove the nmpc abreviation introduction here because it is not used in the abstrac later and you anyway need to introduce it in the main text}
With accelerations of up to 3.5g and speeds over 100 km/h, trajectories generated by the proposed method yield similar or even smaller tracking errors than the trajectories generated for a full multirotor model.
% We will open-source the method.
}
% \robert{the last sentence can be better integrated not to be disjoined} \matej{is it fine if i put it before, like so?}

% \TODO{info about max speed and max thrust used during the experiments}
\end{abstract}
\begin{IEEEkeywords}
    Aerial Systems: Applications; Motion and Path Planning
\end{IEEEkeywords}

\vspace{-0.7em}
\section*{Supplementary Material}
{\footnotesize
\vspace{-0.3em}
\noindent \textbf{Video:} \url{https://youtu.be/wArd536Amro}\\
\noindent \textbf{Code:} \url{https://github.com/ctu-mrs/pmm_uav_planner}
\vspace{-0.7em}
}

%%%%%%%%%%%%%%%%%%%%%%%%%%%%%%%%%%%%%%%%%%%%%%%%%%%%%%%%%%%%%%%%%%%%%%%%%%%%%%%%
\section{Introduction\label{sec:intro}}

\IEEEPARstart{T}{he} problem of finding a minimum-time trajectory for a multirotor Unmanned Aerial Vehicle (UAV) over multiple waypoints holds significant importance for various applications such as search and rescue~\cite{atif2021searchrescue}, inspection and monitoring missions~\cite{ren2019monitoring}, and drone racing~\cite{kaufman2023racing}.
% Efficient trajectory planning is crucial for enabling online replanning onboard the UAV, which is essential for real-time operations in dynamic environments. 
% The ability to compute such trajectories quickly allows for responsive and agile flight, even on less powerful flight computers typically used on small quadrotors.
Finding a trajectory with minimal computational effort is crucial to enable real-time replanning during flight, even on less powerful flight computers used on small multirotors.
% The challenge lies in the fact that this is a nonlinear optimization problem and achieving a truly time-optimal trajectory, as demonstrated in previous research, requires many hours~\cite{Foehn2021TimeOptimal} of computation even for a small number of waypoints, or at least several seconds to achieve near-time-optimality~\cite{fork2023euclidean}.
% The challenge lies in the fact that a nonlinear optimization problem needs to be solved onboard in real-time. 
The challenge lies in the need to solve a nonlinear optimization problem onboard in real-time.
% Achieving a truly time-optimal trajectory, as demonstrated in previous research, requires many hours~\cite{Foehn2021TimeOptimal} of computation, or at least several seconds to achieve near-time-optimality~\cite{fork2023euclidean}.
Achieving a truly time-optimal trajectory, as demonstrated in previous research, requires many hours~\cite{Foehn2021TimeOptimal} of computation, \changed{or at least several seconds~\cite{zhou2023EfficientAR ,fork2023euclidean}.}

Existing works on online trajectory planning often utilize polynomial~\cite{Richter20160423} or  B-spline~\cite{Zhou2021} trajectory representations to compute smooth trajectories efficiently; however, they do not yield minimum-time trajectories due to their smoothness.
Time-optimal trajectories have been found for a full dynamical model of a multirotor using computationally demanding methods~\cite{Foehn2021TimeOptimal,Penicka2022MinTimeCluttered}, or with sampling-based methods~\cite{Foehn2022AlphaPilot},~\cite{Romero2022PMMReplanningMPCC} that solve the task for a double integrator acceleration-limited point-mass model (PMM).
The PMM models only the translational motion of the UAV while ignoring rotational dynamics.
Yet PMM allows for efficient trajectory planning suitable for real-time applications.
However, even with this model, the sampling-based methods could compute trajectories with only a limited number of waypoints in real time.
Additionally, existing approaches utilizing PMM apply per-axis acceleration constraints, which restrict the UAV's ability \changed{to utilize its maximum collective thrust}. %to exploit its actuation potential fully.
Moreover, for distant waypoints, PMM trajectories reaching high speeds may result in large tracking errors without accurate drag modeling, an issue that is yet to be addressed in PMM planning.

% \begin{itemize}
%     \item \cite{Romero2022PMMReplanningMPCC} - summarize
%     \item \cite{Foehn2021TimeOptimal} - summarize
%     \item frontier: generating a trackable trajectory in real-time
% \end{itemize}
% Why hasn’t the problem been solved? What is the stumbling block?
% \begin{itemize}
%     \item Challenging problem of multi-waypoint trajectory generation for a complex non-linear dynamic model of quadrotor.
%     \item Solved by shooting methods within hours in~\cite{Foehn2021TimeOptimal} and within minutes or seconds in~\cite{citeclocation} using collocation.
%     \item TODO
% \end{itemize}
% \item What does our paper contribute?
% \begin{itemize}
%     \item Real-time-capable multi-waypoint minimum-time PMM trajectory generation including crucial real-world constraints.
%     \item Open source implementation.
% \end{itemize}
% \item What is the key idea? What is the magic trick? What is the new insight or technique that enables us to advance the frontier?
% \begin{itemize}
%     \item constraint acceleration norm instead of the per-axis acceleration
%     \item constraint velocity to reflect real-world constraints
%     \item Directly considering gravity to exploit the full potential of collective acceleration produced by the propellers instead of conservative acceleration limits to account for gravity.
%     \item having drag-induced acceleration in the point-mass model
%     \item iterative gradient-method-based trajectory optimization with defined feasibility constraints
% \end{itemize}

\begin{figure}[t]
    \centering
    % \begin{minipage}[b]{0.49\columnwidth}
    %     \centering
    %     \includegraphics[width=\columnwidth, height=0.6\columnwidth]{fig/long_expo.png}
    % \end{minipage}
    % \hfill
    % \begin{minipage}[b]{0.49\columnwidth}
    %     \centering
    %     \includegraphics[width=\columnwidth, height=0.6\columnwidth]{fig/uav.jpg}
    % \end{minipage}
    % \vfill
    % \begin{minipage}[b]{\columnwidth}
        % \centering
        % \includegraphics[width=\columnwidth]{fig/intro_fig_dark.pdf}
        % \includegraphics[width=\columnwidth]{fig/intro_fig_dark_drone.pdf}
        % \includegraphics[width=\columnwidth]{fig/intro_fig_white.pdf}
        % \includegraphics[width=\columnwidth]{fig/title5_c.pdf}
        \includegraphics[width=\columnwidth]{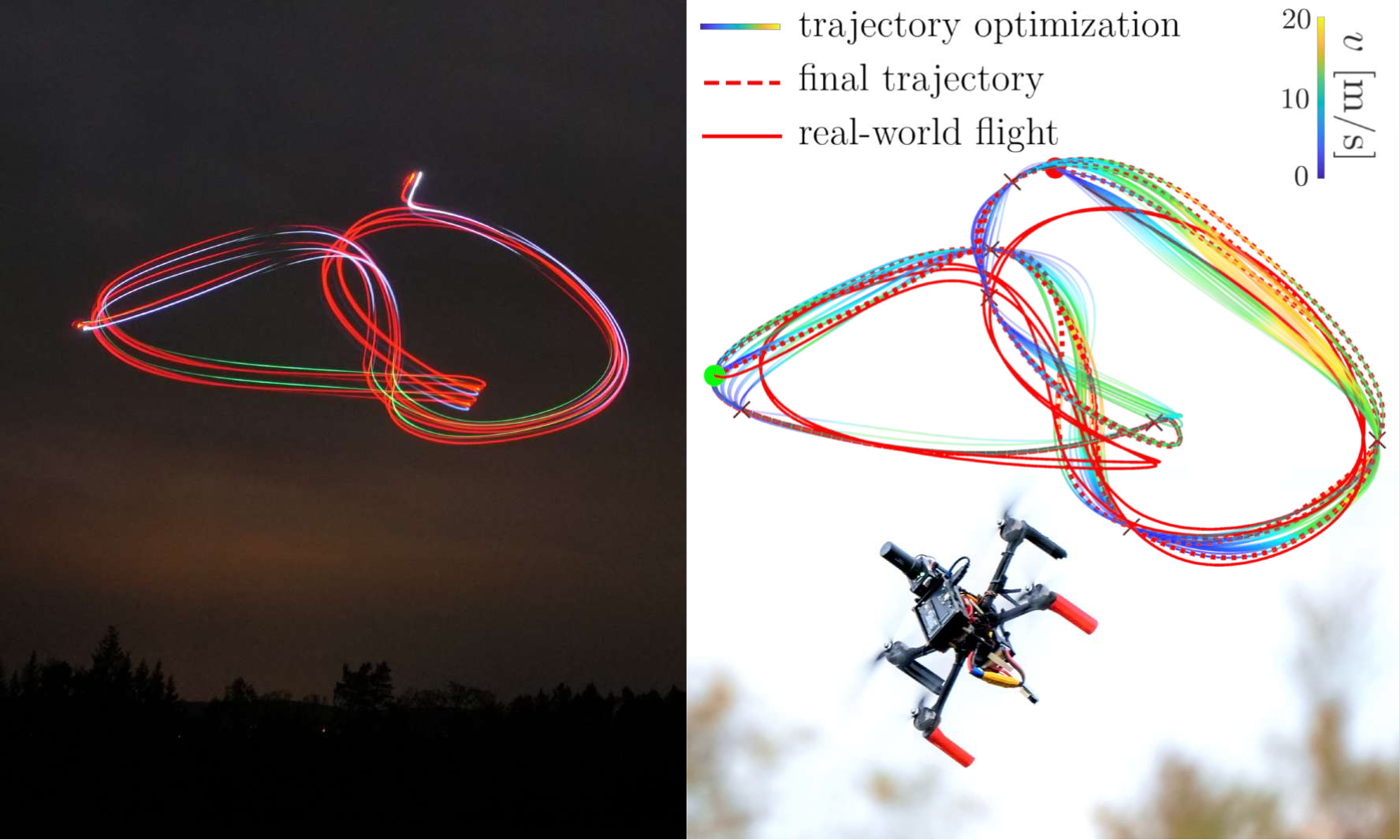}
        % \vspace{-0.5cm}
        % \includegraphics[width=\columnwidth]{fig/opt_traj_viz_v1_edited.pdf}
    % \end{minipage}
    \vspace{-0.7cm}
    \caption{
     Long exposure shot of real-world flight (left) and used UAV together with the visualization of trajectory optimization process (right).
     % \TODO{it is necessary to add reference to the fig1 in the intro!!!}
     % \krystof{The trajectory optimization figure is quite unreadable IMO. I would adjusting the sub-picture ratio (split the figure 50/50 vertically, left still shot, right VO visualization + drone), revert the VO figure to the less transparent version, enlarge it, and, if necessary, even remove the colorbar. I would also put it on a whiter background, not blue, as the plot blends in with it. I think the opptimization process visualization is essential here so that reader gets an idea what0 is happening and it should be clearly visible. What do you think?}
     % \robert{I agree with Krystof, 50/50 would be nicer, also less transparent plot background..., color bar up to you guys}
    \label{fig:intro}}
    \vspace{-0.8cm}
\end{figure}

% To this end, we introduce several enhancements of a computationally efficient minimum-time point-mass model, to address its limitations.
To address these limitations, we present a novel approach, which, within milliseconds, is able to generate minimum-time trajectories over multiple waypoints that can be tracked by a Nonlinear Model Predictive Controller (NMPC) with minimal tracking error.
% \robert{ta veta nad chce prepsat tak aby to neznelo ze jsme neco vylepsili ale tak ze jsme udelali novou vec!!!! slovickareni, ale strasne dulezite. Nesmi to smrdet iterativni praci.}
% Firstly, we implement a Limited Thrust Decomposition algorithm, which iteratively redistributes available acceleration norm constraint into corresponding axis, such that total thrust used is maximized at all times.
We constrain the acceleration norm, optionally also velocity, rather than limiting them per-axis.
% allowing for more agile trajectories that reflect the quadrotor's physical capabilities.
Constraining the norm enables us to use the \changed{maximum collective thrust} of the drone and generate trajectories that are over 20\% faster, compared to the existing works that use multiple, but non-flexible per-axis constraints~\cite{Meyer2023}.
% \robert{constraining norm means that we can use the full actuation potencial of the drone compared to the existing works that use single-axis constraints [cite top uav and similar]}
We also directly consider gravitational force to fully utilize the collective acceleration produced by the propellers, avoiding conservative acceleration limits. % that could be used to account for gravity.
% \robert{ok maybe here cite the works that use those conservative acceleration limits}
% We further refine the model by incorporating simplified drag modeling, which enhances trajectory tracking accuracy, especially at high speeds where aerodynamic drag becomes significant.
% To the best of our knowledge, this is the first work that incorporates drag modeling into the point-mass model, which is of utmost importance if we consider planning between distant waypoints where the minimum-time trajectory reaches extreme velocities, resulting in a massive drag force which is completely disregarded and would result in tracking failure.
The model is further refined by incorporating a linear drag model.
% , which enhances trajectory tracking accuracy, especially at high speeds where aerodynamic drag becomes significant.
This is crucial when planning trajectories between distant waypoints, as minimum-time trajectories can reach extreme velocities, resulting in significant drag forces that, if disregarded, could lead to large tracking errors or even tracking failures. 
To the best of our knowledge, this is the first work to incorporate drag modeling into the PMM. 
% We demonstrate the critical impact of modeling drag-induced acceleration at high speeds, considering gravity, constraining the acceleration norm instead of per-axis acceleration, and distributing acceleration among axes, through an ablation study.
The critical impact of described principles is demonstrated through an ablation study.
% \robert{stress out more the drag as it is important if we consider planning between far waypoints where the min-time trajectory speeds can be large thus the drag could be super large which would mean without modeling drag the tracking of such a trajectory would fail}
% We also introduce a novel gradient-based algorithm for multi-waypoint trajectory optimization, visualized in \reffig{fig:intro}, which optimizes apriori unknown velocities at the waypoints and converges to optimal solutions 100 times faster than the sampling-based approach~\cite{Romero2022PMMReplanningMPCC}.
\changed{We further introduce a novel gradient-based algorithm for multi-waypoint trajectory optimization, visualized in \reffig{fig:intro}, which optimizes apriori unknown velocities at the waypoints and converges to optimal solutions 100 times faster than the sampling-based approach~\cite{Romero2022PMMReplanningMPCC}.}
This is achieved by deriving the key objective function and optimization constraints necessary for fast and successful convergence, which are found in closed form.
% \robert{in the paper we spent a lot of time on how the GD works, the limits, etc. we should sell those things here as well!! e.g. what is the most crutial part/magic tric of the velocity GD that made it work? we should mention it here....} \matej{Could you write something for this @Krystof? It is your expertise :)}
% Finally, we implement an iterative gradient-based optimization method with defined feasibility constraints for a multi-waypoint trajectory generation, capable of quickly converging to optimal solutions.
% We verify our approach in real-flight outdoor experiments showing that trajectories produced within few milliseconds by our method, with maximum thrust of up to 3.5g and velocities reaching over 100~km/h, can be tracked using Nonlinear Model Predictive Controller~(NMPC) with similar or even smaller tracking errors compared to the minimum-time trajectories found within several hours for a full quadrotor model~\cite{Foehn2021TimeOptimal}. 
% We verify our approach in real-flight outdoor experiments (see \reffig{fig:intro}) by tracking trajectories with maximum thrust of up to 3.5g and velocities reaching over 100~km/h using a Nonlinear Model Predictive Controller~(NMPC). 
\changed{We verify our approach in real-flight outdoor experiments (see \reffig{fig:intro}) by tracking trajectories with maximum thrust of up to 3.5g and velocities reaching over 100~km/h using a NMPC.}
Trajectories generated by our method within few milliseconds resulted in similar or even smaller tracking errors than the time-optimal trajectories requiring several hours of computation for a full quadrotor model~\cite{Foehn2021TimeOptimal}. 
\change{We have open-sourced} the method to be used by the community.
\section{Related Work\label{sec:related}}

% \subsection{Trajectory generation \label{sec:related_trajectory_generation}}

% Several different concepts have been used in the field of robotics when planning a trajectory for a UAV.
% For the application of online trajectory replanning for agile flight, computational time and time optimality of the proposed approaches must be considered.
% Several different concepts have been used for trajectory planning for UAVs in the field of robotics.
% Both computational time and time optimality are of critical importance for online trajectory replanning in agile flight.
Several different methods have been used for trajectory planning for agile UAV flight where both computational time and time optimality are of critical importance.
%% Polynomial
In \cite{Richter20160423},\cite{Zhou2019}, \cite{Zhou2021}, the authors utilize the concept of representing a UAV trajectory by continuous-time polynomials to achieve real-time capability.
This representation is enabled by the differential flatness property \cite{Mellinger2011} of a multirotor UAV, where all UAV states and inputs can be expressed using flat outputs and their finite number of derivatives.
Obtaining derivatives of polynomials is computationally effective, which allows for fast computational times.
Quadratic programming is used in~\cite{Richter20160423} to find a polynomial trajectory, which is then optimized with respect to minimal snap and trajectory duration.
Polynomial interpolation using piece-wise polynomial function called B-spline~\cite{Zhou2019} followed by an optimization process combining multiple criteria to achieve perception-aware trajectory is used in~\cite{Zhou2021}.
However, in both cases, the inherent smoothness of polynomials prevents rendering time-optimal policies with rapid input changes.
Direct collocation methods~\cite{fork2023euclidean} achieve near-optimal performance by utilizing polynomials to approximate input and state dynamics, with computation times in the range of tens of seconds. 
% \changed{Recently published approaches~\cite{zhou2023EfficientAR, qin2024timeoptimal} have achieved reduction in the computational time to just seconds.}
A recently published approach~\cite{qin2024timeoptimal} has reduced the computational time to seconds, but achieve only close-to-time-optimal results.

%% Time-optimal
% Time-optimal trajectory planning for a full dynamical model of a multirotor has been introduced in~\cite{Foehn2021TimeOptimal}, where a complex optimization process is proposed, within which a time allocation and a full state are assigned for each of the trajectory waypoints using a discretized dynamic model of the multirotor.
% \changed{Authors in~\cite{hehn2012performancce} plan time-optimal maneuvers in 2D.}
\changed{Time optimal maneuvers were planned in~\cite{hehn2012performancce}, but only in 2D.}
\changed{Time-optimal 3D trajectory planning for a full dynamical model of a multirotor has been introduced in~\cite{Foehn2021TimeOptimal}}, where a time allocation and a full state are assigned for all waypoints using a discretized dynamic model of the multirotor within a complex optimization process.
The authors also address the coupling of linear and rotational acceleration given by the multirotors' design, which is not always addressed in state-of-the-art trajectory planning methods.
Due to the complexity of the proposed approach, the computational time can take several hours, depending on the number of waypoints.
\changed{Yet, authors in~\cite{zhou2023EfficientAR} achieved reduction in runtime to just seconds.}

%% Point-mass model
% To reduce the complexity of trajectory planning for a non-linear and high-dimensional model of a multirotor \cite{Allen20160104}, the approaches \cite{Allen20160104}, \cite{Meyer2022}, \cite{Meyer2023}, \cite{Foehn2022AlphaPilot}, \cite{Penicka2022MinTimeCluttered}, and \cite{Romero2022PMMReplanningMPCC} use a point-mass model (PMM) approximation to generate trajectory, which can then be used to guide recent control methods \cite{Romero2022MPCC} or further trajectory planning for full dynamical model \cite{Allen20160104}, \cite{Foehn2022AlphaPilot}.
\changed{To reduce the complexity of trajectory planning for a non-linear and high-dimensional model of a multirotor \cite{Allen20160104}, the approaches \cite{Allen20160104}, \cite{Meyer2022}, \cite{Meyer2023}, \cite{Foehn2022AlphaPilot}, \cite{Penicka2022MinTimeCluttered}, and \cite{Romero2022PMMReplanningMPCC} use a PMM approximation to generate trajectory, which can then be used to guide recent control methods \cite{Romero2022MPCC} or further trajectory planning for full dynamical model \cite{Allen20160104}, \cite{Foehn2022AlphaPilot}.}
Pontryagin's maximum principle \cite{Bertsekas2007} is used to achieve a time-optimal PMM trajectory, which results in a bang-bang policy in the case of limited acceleration trajectories \cite{Allen20160104} and a bang-singular-bang policy for limited acceleration and velocity \cite{Meyer2022}.
%% Bang-singular-bang
\changed{In \cite{Meyer2022}, bang-singular-bang policy PMM trajectory generation was used under the Kinematic Orienteering Problem. %, which also addresses the ambiguity in the states of the waypoints.
% Additional synchronization patterns are introduced to solve the trajectory axis synchronization task.
% The enhancement of their methodology was demonstrated in \cite{Meyer2023}, where the authors introduced four options for distributing maximum velocity and acceleration between each trajectory axis.
The enhancement of their methodology was demonstrated in \cite{Meyer2023}, where four per-axis cases for velocity and acceleration constraints were introduced.
% However, the assumption of known full states at all waypoints makes the method unsuitable for multi-waypoint trajectory optimization.
However, the assumption of known full states at all waypoints makes the method unsuitable for multi-waypoint trajectories.
% The authors of~\cite{Foehn2022AlphaPilot} use bang-singular-bang trajectories with axis synchronization managed by scaling single-axis time-optimal trajectories. 
% Velocity ambiguity for multi-waypoint trajectories is resolved by sampling different velocity vectors at each waypoint.
The authors of~\cite{Foehn2022AlphaPilot} also use bang-singular-bang trajectories but solve the ambiguity for multi-waypoint trajectories by sampling different velocity vectors at each waypoint.
The minimum duration trajectory is then fitted with a polynomial, retaining thus the previously mentioned drawbacks.}
%% Bang-bang
Limited acceleration PMM trajectory planning was implemented in \cite{Allen20160104} within an obstacle avoidance framework.
% Axis synchronization was not addressed, as these trajectories served only as approximate guiding trajectories for subsequent spline-based planning.
\changed{However, no axis synchronization was addressed.}
The ability of modern control methods to track infeasible but differentiable paths was demonstrated in \cite{Penicka2022MinTimeCluttered}, \cite{Romero2022PMMReplanningMPCC}, where a Model Predictive Contouring Control \cite{Romero2022MPCC} method is used to track PMM bang-bang trajectories.
% Similar to \cite{Foehn2022AlphaPilot}, the ambiguity in velocities for multi-waypoint trajectories is solved by generating gradually adjusted velocity samples from inside a cone spreading from the corresponding waypoint towards the next to increase efficiency; the samples that result in the shortest trajectory duration are selected.
Similar to \cite{Foehn2022AlphaPilot}, the ambiguity in velocities for multi-waypoint trajectories is solved by generating gradually adjusted velocity samples from inside a cone; the samples that result in the shortest trajectory duration are selected.
% Throughout the sampling process, the sampling cone is refocused based on the best sample currently available, i.e., the state that results in the shortest trajectory duration.
% In \cite{Penicka2022MinTimeCluttered}, a gradient descent method on a sphere was added to optimize the thrust acceleration decomposition in a bang-bang policy trajectory and solve the problem of coupled rotational and linear acceleration of a multirotor, obtaining thus the minimum-time trajectories.
In \cite{Penicka2022MinTimeCluttered}, a gradient descent method on a sphere was added to optimize the thrust decomposition and solve the problem of coupled rotational and linear acceleration of a multirotor, obtaining thus the minimum-time trajectories.
% However, as the computational complexity of the sampling method grows quadratically with the number of velocity samples per waypoint, a receding horizon approach had to be used in a real-time replanning application \cite{Romero2022PMMReplanningMPCC}, where the trajectory was generated only for few subsequent waypoints.
\changed{However, as the computational complexity of this method grows quadratically with the number of samples per waypoint, a receding horizon approach had to be used in a real-time replanning application \cite{Romero2022PMMReplanningMPCC}, where the trajectory was generated only for few subsequent waypoints.}
Our method can converge to optimal solutions 100 times faster thanks to continuous velocity optimization.

% \TODO{decide where to put this...}
\changed{While PMM effectively reduces the complexity of planning by neglecting rotational dynamics including the heading, theoretically optimal trajectories are not feasible since they require infinite angular acceleration to change the direction of thrust instantaneously. 
% However, the infeasibility can be addressed by employing a NMPC, which takes into account the missing vehicle's rotational dynamics and physical constraints.
% As result, NMPC is able to track infeasible trajectories with similar tracking errors as trajectories planned for a full dynamical model.
% While PMM simplifies planning by neglecting rotational dynamics, its theoretically optimal trajectories are infeasible, requiring infinite angular acceleration for instant thrust changes. 
The infeasibility and heading ambiguity can be resolved by using NMPC, which accounts for the vehicle's rotational dynamics and physical constraints, and can track infeasible trajectories with errors comparable to trajectories planned for full dynamical model.}

% \section{Problem Statement\label{sec:problem}}

% \TODO{Not sure if we need this section, please think of it if yes or not, consider most related works if they do have it or not, maybe we can join this section rather with the methodology....}

% \krystof{Probably depends on how extensively we want to include the NMPC, are we going to cite a separate article or introduce it here? K.}
% \robert{Hmmmm gut point, I would do it now without the NMPC and focus only on the replanning and after we converge to a certain paper we can reevaluate if we want to add the NMPC part... The thing is in the NMPC there is nothing novel (yet) so it would be only some math to show off, on the other hand it would make it the approach more self-contained because without the NMPC it won't fly with the trajectories as they are bang-bang. So, lets keep both as options.}

\section{Methodology\label{sec:method}}

%% --------------------------------------------------------------
%% | Trajectory generation                                      |
%% --------------------------------------------------------------
\subsection{Time-Optimal Point-Mass Model Trajectory Planning}
\label{sect:pmm}

% \subsubsection{Point-mass model trajectory}
%% BANG BANG
% For trajectory segment of a point-mass model (PMM) with limited acceleration resulting in a bang-bang policy, we define the start and end state as positions \mbox{$ \mathbf{p}_0, \mathbf{p}_2 \in \mathbb{R}^n $} and velocities \mbox{$ \mathbf{v}_0, \mathbf{v}_2 \in \mathbb{R}^n $}, where $n$ is the dimensionality of the space in which the trajectory is computed; $n=3$ for our application.
\changed{For a PMM trajectory segment $\Pi$ with limited acceleration resulting in a bang-bang policy, we define the start and end state as positions \mbox{$ \mathbf{p}_0, \mathbf{p}_2 \in \mathbb{R}^n $} and velocities \mbox{$ \mathbf{v}_0, \mathbf{v}_2 \in \mathbb{R}^n $}, where $n$ is the dimensionality of the space in which the trajectory is computed; $n=3$ for our application.}
% The control input is the acceleration $ \mathbf{a} \in \mathbb{R}^n$, which is bounded by the per-axis acceleration limits $ \mathbf{a}_{\text{min}} $,~$ \mathbf{a}_{\text{max}} \in \mathbb{R}^n \setminus \{0\} $, where the $j$-th component of $\mathbf{a}$ holds $ {a}^{j}_{\text{min}} \le {a}^{j} \le {a}^{j}_{\text{max}} $, $ j = 1,...,n$. 
\changed{The control input is the acceleration $ \mathbf{a} \in \mathbb{R}^n$ in the world frame, which is bounded by the per-axis acceleration limits $ \mathbf{a}_{\text{min}} $,~$ \mathbf{a}_{\text{max}} \in \mathbb{R}^n \setminus \{0\} $, where the $j$-th component of $\mathbf{a}$ holds $ {a}^{j}_{\text{min}} \le {a}^{j} \le {a}^{j}_{\text{max}} $, $ j = 1,...,n$. }
We also consider $ \mathbf{a} \neq \mathbf{0} $ for a non-zero trajectory with $ \mathbf{p}_0 \neq \mathbf{p}_2 $.
Without loss of generality, we define a single-axis trajectory segment first.
Given the Pontryagin's maximum principle \cite{Bertsekas2007}, a time-optimal PMM trajectory segment can be described as:
\begin{equation}\label{eq:pmm_1seg}
	\begin{aligned}
		p_1 &= p_0 + v_0 t_1 + \dfrac{1}{2} a_1 t_1^2, \qquad
		v_1 = v_0 + a_1 t_1,\\
		p_2 &= p_1 + v_1 t_2 + \dfrac{1}{2} a_2 t_2^2, \qquad
		v_2 = v_1 + a_2 t_2,
	\end{aligned}
\end{equation}
% where $ p_0, v_0 \in \mathbb{R} $ are the start position and velocity, $ p_2, v_2~\in~\mathbb{R} $ are the end position and velocity, $ t_1, t_2 \in \mathbb{R} $ are the time durations of the corresponding sub-segments of the trajectory and
where $ p_0, v_0 \in \mathbb{R} $ and $ p_2, v_2\in\mathbb{R} $ are the start and end states, $ t_1, t_2 \in \mathbb{R} $ are the time durations of the corresponding sub-segments of the trajectory and
\begin{equation}\label{eq:pmm_acc_limits}
	a_i \in \{a_{\text{min}}, a_{\text{max}}\}, \; i = 1,2, \; a_i \in \mathbb{R},
\end{equation}
are the optimal control inputs.
We omit the case where \mbox{$ a_1=a_2 $} as this is equivalent to $ t_1 = 0 $ or $ t_2 = 0 $, which leaves us with two combinations of possible values for $ a_1 $ and $ a_2 $.
Given the start state, end state, and limits on the acceleration, the unknowns are the sub-segment durations $ t_1, t_2 $, velocity $ v_1 $ and position $ p_1 $ at time $ t = t_1 $.
It can be shown that analytical solutions to the equations~\refeq{eq:pmm_1seg} exist and that there are four solutions in total; for each combination of $ a_1 $ and $ a_2 $ values, there are two.
From those, the final solution is chosen such that all unknowns are real numbers, and the real-world condition of non-negative times $ t_1 \geq 0 $ and $ t_2 \geq 0 $ is satisfied.
% \krystof{Replaced $ t_1 > 0 $ with $t_1 \geq 0 $ in order to be non-negative, please doublecheck.}
The total trajectory duration of the selected time-optimal solution is
\begin{equation}\label{eq:pmm_tr_time}
	 T_{\changed{\text{ax}}} = t_1 + t_2.
\end{equation}

%% BANG-SINGULAR-BANG
% The bang-bang policy is impractical for applications with larger distances between waypoints, as the resulting large velocities are infeasible due to the physical limits of the UAV.
The bang-bang policy is impractical for longer distances between waypoints, as the resulting high velocities induce drag forces greater than the UAV can counteract.
% The bang-bang policy is impractical for longer distances between waypoints, as the induced drag forces can yeld the high velocities unreachable.
Therefore, we introduce an optional velocity limit $ {v}_{m} > 0, \; \lvert v(t) \rvert  \leq v_m \; \text{for} \; t\in[0, T_{\changed{\text{ax}}}]$.
The time-optimal control inputs for bang-singular-bang policy are
\begin{equation}\label{eq:pmm_bsb_acc_limits}
	a_i \in \{a_{\text{min}}, 0, a_{\text{max}}\}, \; i = 1,2,3, \; a_i \in \mathbb{R},
\end{equation}
with notation analogous to \refeq{eq:pmm_1seg}. 
If the speed limit is reached, the value of $v_1$ is known and equal to the maximum velocity $\pm v_{m}$. Then, an additional equation
% \begin{equation}
    $p_\sigma = p_1 + v_{m} t_\sigma$,
% \end{equation}
% to equations~\refeq{eq:pmm_1seg} needs to be solved.
% This equation involves two new variables: the duration $t_\sigma$ during which the acceleration $a_i=0$, and the position $p_\sigma$ at which the zero acceleration segment ends.
is added to the system~\refeq{eq:pmm_1seg}, where the two new unknowns are the duration $t_\sigma$ during which the acceleration $a_i=0$ is applied, and the position $p_\sigma$ at which the zero acceleration segment ends.
It can be shown that the solution for all unknowns can be found in closed form.
% , and we select the final solution according to the same criteria.
The total trajectory duration for the bang-singular-bang policy trajectory segment is then
\begin{equation}\label{eq:pmm_bsb_tr_time}
    T_{\changed{\text{ax}}} = t_1 + t_\sigma + t_2.
\end{equation}

\subsection{Trajectory Axis Synchronization}
\label{sect:ax_sync}
% Generalizing the introduced concept of PMM trajectory to a multidimensional trajectory, the issue of axis synchronization arises, where the time-optimal trajectory durations $ T_{\text{ax}_1}, \ldots, T_{\text{ax}_n} $ (as defined in equation \refeq{eq:pmm_tr_time} or \refeq{eq:pmm_bsb_tr_time}) for all $n$ axes need to be synchronized to a common duration $T_{s}$.
The issue of generalizing the introduced concept of PMM trajectory to a multidimensional trajectory leads to the problem of axis synchronization. 
The time-optimal trajectory durations $ T_{\text{ax}_1}, \ldots, T_{\text{ax}_n} $ (as defined in equation \refeq{eq:pmm_tr_time} or \refeq{eq:pmm_bsb_tr_time}) for all $n$ axes need to be synchronized to a common duration $T_{\text{s}}$.
We follow the approaches \cite{Foehn2022AlphaPilot} and \cite{Romero2022PMMReplanningMPCC}, where $ T_{\text{s}} $ is set to the duration of the slowest per-axis trajectory and acceleration scaling factor \mbox{$\gamma\in(0,1]$}, different for each axis, is used for scaling down the accelerations in the rest of the axes to prolong their duration.
By substituting $ \gamma_{i} a_i $ for both $a_1$ and $a_2$ in \refeq{eq:pmm_1seg} and adding the condition of the set trajectory duration $T_{\text{s}} = t_1 + t_2 $, one can obtain the solution for all unknowns in closed form.
% We again choose the solution where all unknowns are real numbers, the real-world condition of non-negative time is met, and $ \gamma$ lies in the defined interval.
We select the solution where all unknowns are real numbers, time is non-negative, and $ \gamma$ is within the defined interval.
However, such a solution does not necessarily exist as described in \cite{Meyer2023} since the correct position has to be integrated using the scaled accelerations in the given time.
% If no solution was found, the new feasible synchronization duration $ T_{\text{s}_{\text{new}}} $ is selected by computing the shortest-duration feasible \mbox{$T_{\text{tr}}>T_{\text{s}}$} for the given axis.
% It can be shown by studying the feasibility intervals of a given trajectory based on the scaling factor $\gamma$, that the next feasible trajectory with the shortest duration can be computed by solving equations~\refeq{eq:pmm_1seg} with swapped acceleration values $ a_{1_{\text{new}}} := a_{2_{\text{old}}}, \, a_{2_{\text{new}}} := a_{1_{\text{old}}} $.
\changed{
If no solution was found, the new feasible synchronization duration $ T_{\text{s}_{\text{new}}} $ is computed.
This is done by setting $\gamma=1$ and solving \refeq{eq:pmm_1seg} for the given axis.
The feasible solution, for which $T_{\changed{\text{ax}}}$ is the shortest and \mbox{$T_{\changed{\text{ax}}}>T_{\text{s}}$} holds, is selected as the next synchronization duration.
This is equivalent to obtaining the edge value of the nearest interval of feasible $T_{\changed{\text{ax}}}$ values, which are directly determined by the edge values of $\gamma\in(0,1]$.
We solve only for $\gamma=1$ as $\gamma\rightarrow 0 $ yields $T_{\changed{\text{ax}}} \rightarrow \infty$.
% To our knowledge, we are the first to present this method for axis synchronization of a limited-acceleration PMM trajectory.
}
%Given the $ T_{\text{s}_{\text{new}}} $, the synchronization is run for all the remaining axes as described above.
\changed{Given the $ T_{\text{s}_{\text{new}}} $, the synchronization is performed for all the remaining axes as described above.}
We refer to the above-described method for a synchronized one-segment multidimensional PMM trajectory as $\mathbf{pmmTraj3D}$ in \refalg{alg:thrust_decomposition} and \refalg{alg:velocity_optimization}.
% \changed{A scheme depicting our approach is shown in \reffig{fig:method_scheme}.}

\subsection{Limited Thrust Decomposition (LTD)}
\label{sect:ltd}
% Until now, we used per-axis acceleration limits.
% To allow for the properties of multirotor UAV flight, where the three-axis accelerations are coupled with the limited thrust vector of the multirotor, aerodynamic drag is induced as a consequence of fast flight and gravity plays a significant role, we are introducing the following approach.
% Given a maximal thrust force $f_{\text{max}}$ the motors of a multirotor with mass $ m $ can produce, we compute its limited thrust acceleration limit as \mbox{$a_{{\text{max}}}^T={f_{\text{max}}}/{m}$}.
% However, to account for the unique properties of multirotor UAV flight, including the coupling of three-axis accelerations with the limited thrust vector, the aerodynamic drag from fast flight, and the significant role of gravity, we are introducing a new approach. 
Until now, we used per-axis acceleration limits. 
Yet, the unique property of multirotors is the coupling of the three-axis accelerations with the collective thrust vector produced by the motors with propellers. 
Additionally, external forces such as gravity and aerodynamic drag induced by fast flight need to be considered.
To this end, we propose the following approach.
% Given a maximum collective thrust force $f_{\text{max}}$ that the motors of a multirotor with mass $m$ can produce, we compute its thrust acceleration limit as \mbox{$a_{\text{max}}^T={f_{\text{max}}}/{m}$}.
\changed{Given a maximum collective thrust force $f_{\text{max}}$ that the motors of a multirotor with mass $m$ can produce, we compute its thrust acceleration $\mathbf{a}^{T}$ limit as \mbox{$a_{\text{max}}^T={f_{\text{max}}}/{m}$}.}
% The body acceleration $\mathbf{a}^B$ of a trajectory must then for all times $ t \in [0, T_{\text{tr}}] $ satisfy the constraint
\changed{The acceleration $\mathbf{a}$ of a trajectory must then for all times $ t \in [0, T_{{\text{s}}}] $ satisfy the constraint}
\begin{equation}\label{eq:ltd_body_thrust}
	% \mathbf{a}^T = \mathbf{a}^B - \mathbf{d} - \mathbf{g},
	\lVert \mathbf{a}^T(t) \rVert = \lVert \changed{\mathbf{a}}(t) - \mathbf{d}(t) - \mathbf{g} \rVert \leq {a_{{\text{max}}}^T},
\end{equation}scheme
where $ \mathbf{g}=[0,0,-g]^\top $ and \changed{$\mathbf{d}=[-d^x, -d^y, -d^z]^\top$} are gravitational and drag acceleration vectors, respectively.

\changed{Rather than deriving a complex optimization approach to solve the non-linear constraint \refeq{eq:ltd_body_thrust}, and to minimize the computational time, we build on the work \cite{Penicka2022MinTimeCluttered} and iteratively approximate an optimal distribution of accelerations 
% \begin{equation}\label{eq:ltd_acc_fcn}
% 	\mathbf{a}^B(t) = [a_x^B(t), a_y^B(t), a_z^B(t)]^\top, \; t\in [0,T_{\text{tr}}]
% \end{equation}
to minimize the $ T_{\text{s}} $ of a PMM trajectory segment.}
% For a single three-dimensional bang-bang policy trajectory segment with per-axis trajectory equations \refeq{eq:pmm_1seg}, the problem can be formulated as \refeq{eq:ltd_problem}.

% \begin{equation}\label{eq:ltd_problem}
% 	\begin{aligned}
% 	\{\mathbf{a}_1^{B*}, \mathbf{a}_2^{B*}\} &= \arg\min_{\mathbf{a}_1^B, \mathbf{a}_2^B \in \mathbb{R}^3} T_{\text{tr}}(\mathbf{a}_1^B, \mathbf{a}_2^B) \\
% 	\text{s.t.} \;\;
% 		\mathbf{a}_1^B &= [a_{1_x}^B, a_{1_y}^B, a_{1_z}^B]^\top, \\
% 		\mathbf{a}_2^B &= [a_{2_x}^B, a_{2_y}^B, a_{2_z}^B]^\top, \\
% 		{a}_i^B(t) &= {a}_{1_i}^B \; \text{for} \; t \in [0, t_{1_i}], \; \text{for}\; i \in \{x,y,z\}, \\
% 		{a}_i^B(t) &= {a}_{2_i}^B \; \text{for} \; t \in [t_{1_i}, T_{\text{tr}}], \; \text{for}\; i \in \{x,y,z\},\\
%         % \mathbf{d} &= \delta \cdot \mathbf{v}(t), \; \delta<0 \\
% 		\lVert \mathbf{a}^B(t)& - \mathbf{d}(t) - \mathbf{g}\rVert \leq a_{{\text{max}}}^T, \; \text{for} \; t \in [0, T_{\text{tr}}].\\
% 	\end{aligned}
% \end{equation}
\changed{For a single three-dimensional bang-bang policy trajectory segment with per-axis trajectory equations \refeq{eq:pmm_1seg}, the problem can be formulated as
\begin{equation}\label{eq:ltd_problem}
	\begin{aligned}
	\{\mathbf{a}_1^{*}, \mathbf{a}_2^{*}\} &= \arg\min_{\mathbf{a}_1, \mathbf{a}_2 \in \mathbb{R}^3} T_{\changed{\Pi}}(\mathbf{a}_1, \mathbf{a}_2) \\
	\text{s.t.} \;\;
		{a}_i(t) &= {a}_{1}^i \; \text{for} \; t \in [0, t_{1}^i], \; \text{for}\; i \in \{x,y,z\}, \\
		{a}_i(t) &= {a}_{2}^i \; \text{for} \; t \in [t_{1}^i, T_{\changed{\Pi}}], \; \text{for}\; i \in \{x,y,z\},\\
        % \mathbf{d} &= \delta \cdot \mathbf{v}(t), \; \delta<0 \\
		\lVert \mathbf{a}(t)& - \mathbf{d}(t) - \mathbf{g}\rVert \leq a_{{\text{max}}}^T, \; \text{for} \; t \in [0, T_{{\text{s}}}],\\
	\end{aligned}
\end{equation}
% where $\mathbf{a}_1 = [a_{1}^x, a_{1}^y, a_{1}^z]^\top$, $\mathbf{a}_2 = [a_{2}^x, a_{2}^y, a_{2}^z]^\top$, and $T_{\changed{\Pi}}$ is a function mapping the used acceleration vectors to the duration of a synchronized trajectory defined in \refsec{sect:ax_sync}.
where $\mathbf{a}_1 = [a_{1}^x, a_{1}^y, a_{1}^z]^\top$, $\mathbf{a}_2 = [a_{2}^x, a_{2}^y, a_{2}^z]^\top$, and $T_{\changed{\Pi}}$ is a function assigning the duration of a synchronized trajectory defined in \refsec{sect:ax_sync} to the selected accelerations.}

% The duration of the first sub-segment for the $i$-axis $ t_{1_i} $, is referred to as acceleration switch time, and $T_{\text{tr}}$ is the total trajectory duration, solution to \refsec{sect:ax_sync}.
\changed{The duration of the first sub-segment for the $i$-axis $ t_{1}^i $ is referred to as acceleration switch time, and $T_{{\text{s}}}$ is the total trajectory duration, solution to \refsec{sect:ax_sync}.}
% To approximate the solution to \refeq{eq:ltd_problem}, we first decompose the maximal thrust $ a_{{\text{max}}}^T $ into per-axis limits $ \mathbf{a}_{\text{max}} $ and $ \mathbf{a}_{\text{min}} $ by solving~\refeq{eq:ltd_body_thrust} for \mbox{$a_{\text{init}}^B=a_x^B = a_y^B = a_z^B $}.
% To approximate the solution to \refeq{eq:ltd_problem}, we first decompose the maximal thrust $ a_{{\text{max}}}^T $ into per-axis limits $ \mathbf{a}_{\text{max}} $ and $ \mathbf{a}_{\text{min}} $ by solving~\refeq{eq:ltd_body_thrust} for \changed{\mbox{$a_{\text{init}}=a^x = a^y = a^z $}}.
\changed{To approximate the solution to \refeq{eq:ltd_problem}, we first decompose the maximal thrust $ a_{{\text{max}}}^T $ into per-axis limits $ \mathbf{a}_{\text{max}} $ and $ \mathbf{a}_{\text{min}} $ by solving~\refeq{eq:ltd_body_thrust} for \changed{\mbox{$a_{\text{init}}=a^x = a^y = a^z $}} given that $\lVert \mathbf{a}^T \rVert = a_{{\text{max}}}^T$.
We also consider $\mathbf{d} = \mathbf{0}$ for this initialization step as the velocities necessary for drag estimation need to be computed first.}
This can be done in closed form with two solutions, from which we select the positive number.
% The initial body per-axis acceleration limits are computed as
\changed{The initial per-axis acceleration limits are computed as}
% \begin{equation}\label{eq:ltd_acc_limits}
% 		\changed{\mathbf{a}_{\text{max}} = [a_{\text{init}} ,a_{\text{init}} , a_{\text{init}}]^\top}, \qquad
% 		\changed{\mathbf{a}_{\text{min}}} = -\changed{\mathbf{a}_{\text{max}}} + 2\mathbf{g}.
% \end{equation}
\changed{\begin{equation}\label{eq:ltd_acc_limits}
    \begin{aligned}
		\mathbf{a}_{\text{max}} &= \mathbf{a}^{T}_{\text{init}} + \mathbf{g} = [a_{\text{init}} ,a_{\text{init}} , a_{\text{init}}]^\top,\\
	   % \mathbf{a}_{\text{min}} &= -\mathbf{a}^{T}_{\text{init}} - \mathbf{d} + \mathbf{g} = \underbrace{-\mathbf{a}_{\text{max}} + \mathbf{d} + \mathbf{g}}_{= -\mathbf{a}^{T}_{\text{init}}} - \mathbf{d} + \mathbf{g} \\
    %     &= \mathbf{a}_{\text{max}} + 2\mathbf{g},
        % \mathbf{a}_{\text{min}} &= -\mathbf{a}^{T}_{\text{init}} + \mathbf{g} = -\mathbf{a}_{\text{max}} + \mathbf{g} + \mathbf{g} = -\mathbf{a}_{\text{max}} + 2\mathbf{g},
        \mathbf{a}_{\text{min}} &= -\mathbf{a}^{T}_{\text{init}} + \mathbf{g} = -\mathbf{a}_{\text{max}} + 2\mathbf{g},
    \end{aligned}
\end{equation}
where $\mathbf{a}^{T}_{\text{init}}$ is the initial thrust acceleration.
}

Using these values, we compute the multidimensional trajectory as described in Sec.~\ref{sect:ax_sync}.
As synchronizing trajectories are computed using acceleration downscaling, the acceleration norm will generally be smaller than $ a_{\text{max}}^T $ throughout the trajectory segment.
% According to Pontryagin's maximum principle \cite{Bertsekas2007}, the maximal acceleration possible is required to obtain time optimality.
\changed{According to Pontryagin's maximum principle \cite{Bertsekas2007}, the maximal acceleration is required to obtain time optimality.}
For our case, a thrust acceleration vector $ \mathbf{a}^T $ must at some point in time satisfy
\begin{equation}\label{eq:ltd_max_acc_condit}
	\lVert \mathbf{a}^T \rVert = a_{\text{max}}^T.
\end{equation}

% Using the bang-bang policy, we have three different per-axis switch times $ t_{1_i}, i\in \{x, y, z\} $ which lead to four different thrust acceleration vectors $ \mathbf{a}_l^T, \; l\in{1,\dots,4} $ in one segment.
Using the bang-bang policy, we have three different per-axis switch times \changed{$ t_{1}^i, i\in \{x, y, z\} $} which lead to four different thrust acceleration vectors $ \mathbf{a}_l^T, \; l\in{1,\dots,4} $ in one segment.
% Due to the control input structure, the drag vector $\mathbf{d}$ is approximated using a linear drag model used in~\cite{faessler2017flatness}, for the most adverse scenarios for each \changed{$ \mathbf{a}_l$}, which always occurs right before per-axis switch times
\changed{
Using a linear drag model used in~\cite{faessler2017flatness}, drag vector $\mathbf{d}$ is approximated for the most adverse scenarios for each \changed{$ \mathbf{a}_l$}, guaranteed to be right before per-axis switch times, as}
\begin{equation}\label{eq:drag}
     % \mathbf{d}_l = \delta\cdot \mathbf{v}(t_l), \; t_l \in \{t_{1_x}, t_{1_y}, t_{1_z}, T_{\text{tr}}\},
     \mathbf{d}_l = \mathbf{R}(t_l)\mathbf{D}\mathbf{R}^{\top}(t_l)\cdot\mathbf{v}(t_l),
\end{equation}
% where rotation matrix $\mathbf{R}(t)$ can be obtained from current acceleration, $\mathbf{D}=\text{diag}(\delta_x, \delta_y, \delta_z)$ is a diagonal matrix containing drag coefficients for each axis, and \mbox{$t_l \in \{t_{1_x}, t_{1_y}, t_{1_z}, T_{\text{tr}}\}$}.
where rotation matrix $\mathbf{R}(t)$ can be obtained from current acceleration, \changed{$\mathbf{D}=\text{diag}(\delta^x, \delta^y, \delta^z)$} is a diagonal matrix containing drag coefficients for each axis, and \changed{\mbox{$t_l \in \{t_{1}^x, t_{1}^y, t_{1}^z, T_{{\text{s}}}\}$}}.
In these instances, velocity is either maximized during acceleration, resulting in the largest drag acting opposite acceleration, or minimized during deceleration, with the smallest drag aligned in the same direction as the acceleration vector.
% In these instances, the peak velocity values are reached and the largest drag forces act upon the body of the UAV. 
% Since actual drag is guaranteed to be smaller then approximated drag at switch times $t\in[t_{1_x}, t_{1_y}, t_{1_z}]$ we only need to account for two values of the thrust acceleration magnitude in the given trajectory to fulfill constraint~\refeq{eq:ltd_body_thrust}
Thus, over the duration of the trajectory, we observe four local maxima for thrust acceleration
\begin{equation}
    \label{eq:acc_cases}
	% \lVert \mathbf{a}_l^T \rVert \in \left\{\lVert \mathbf{a}_{1}^B-\mathbf{d_1}-\mathbf{g} \rVert,\lVert \mathbf{a}_{2}^B-\mathbf{d_2}-\mathbf{g} \rVert\right\},
    \lVert \mathbf{a}_l^T \rVert = \lVert \changed{\mathbf{a}_{l}}-\changed{\mathbf{d}_l}-\mathbf{g} \rVert, \; l \in \{1, \dots, 4\}.
\end{equation}
% where $ \mathbf{a}_1^B := \mathbf{a}_{l=1}^B $ is the start acceleration, $ \mathbf{a}_2^B := \mathbf{a}_{l=4}^B $ is the end acceleration.
% To find the body per-axis acceleration limits $\changed{\mathbf{a}_{\text{max}}}$ and $ \changed{\mathbf{a}_{\text{min}}}$ which result in satisfying the condition \refeq{eq:ltd_max_acc_condit} for all thrust accelerations $ \mathbf{a}_l^T $, we propose the following iterative approach.
\changed{To find the per-axis acceleration limits $\changed{\mathbf{a}_{\text{max}}}$ and $ \changed{\mathbf{a}_{\text{min}}}$ which result in satisfying the condition \refeq{eq:ltd_max_acc_condit} for all thrust accelerations $ \mathbf{a}_l^T $, we propose the following iterative approach.}
In each step, all $\mathbf{a}_l^B$ are scaled by their unique factors $\beta_l$ to achieve \refeq{eq:ltd_max_acc_condit}.
% In each step, new per-axis acceleration limits $ \mathbf{a}_{\text{min}_{j+1}}^B $ and $ \mathbf{a}_{\text{max}_{j+1}}^B$ are approximated by finding a factor $ \beta \in \mathbb{R} $ with which the current $\mathbf{a}_{1_j}^T$ and $\mathbf{a}_{2_j}^T$ are scaled to achieve \refeq{eq:ltd_max_acc_condit} as follows
% \begin{equation}\label{eq:new_acc_bounds}
		% \mathbf{a}_{\text{L1}_{j+1}}^B = \beta_1 \mathbf{a}_{1_j}^T +\mathbf{d_1} +\mathbf{g}, \quad
		% \mathbf{a}_{\text{L2}_{j+1}}^B = \beta_2 \mathbf{a}_{2_j}^T +\mathbf{d_2} +\mathbf{g}.
  %       \mathbf{a}_{\text{L1}_{j+1}}^B = \beta_1 \mathbf{a}_{1_j}^B, \quad
		% \mathbf{a}_{\text{L2}_{j+1}}^B = \beta_2 \mathbf{a}_{2_j}^B.
% \end{equation}
% The factors $ \beta_1 $ and $ \beta_2 $ are computed by solving the equations
The factors are computed by solving the equation
\begin{equation}\label{eq:ltd_beta2}
    \begin{aligned}
	% {a_{\text{max}}^T}^2 = \left(\beta_i a_{i_{x_j}}^B+d_{i_x}\right)^2+\left(\beta_i a_{i_{y_j}}^B+d_{i_y}\right)^2\\+\left(\beta_i a_{i_{z_j}}^B+d_{i_z}+ g\right)^2, \;\text{for} \; i \in \{1, 2\},
    % {a_{{\text{max}}}^T} = \lVert \beta_i\mathbf{a}_{i_j}^B - \mathbf{d}_{i_j} - \mathbf{g} \rVert, \,\text{for} \, i \in \{1, 2\},
    {a_{{\text{max}}}^T} = \lVert \beta_l\changed{\mathbf{a}_{l}} - \mathbf{d}_{l} - \mathbf{g} \rVert, \, l \in \{1, \dots, 4\},
    \end{aligned}
\end{equation}
% where $ \mathbf{a}_{l} = [a_{l_{x}}^B, a_{l_{y}}^B, a_{l_{z}}^B]^\top $, and which can be solved in closed form.
% We choose a positive real-number value out of the two possible solutions in order to correctly assign the acceleration values.
where \changed{$ \mathbf{a}_{l} = [a_{l}^x, a_{l}^y, a_{l}^z]^\top $}, and which can be solved in closed form.
We choose a positive real-number value out of the two possible solutions in order to correctly assign the acceleration values.
% New body per-axis acceleration limits $\changed{\mathbf{a}_{\text{max}}}$ and $ \changed{\mathbf{a}_{\text{min}}}$ are then determined as
\changed{New per-axis acceleration limits $\changed{\mathbf{a}_{\text{max}}}$ and $ \changed{\mathbf{a}_{\text{min}}}$ are then determined as}
% \begin{equation}\label{eq:new_acc_bounds}
% 	% \begin{aligned}
%  %        a_{\text{max}_{i}}^B = \min\{\beta_l a_{l_{i}}^B \;|\; a_{l_{i}}^B > 0\} &\quad i \in \{x, y, z\} \\ 
% 	% 	a_{\text{min}_{i}}^B = \max\{\beta_l a_{l_{i}}^B \;|\; a_{l_{i}}^B < 0\} &\quad l \in \{1, \dots, 4\}
%  %    \end{aligned}
%  	\begin{tabular}{ccc}
%         $a_{\text{max}_{i}}^B = \min\{\beta_l a_{l_{i}}^B \;|\; a_{l_{i}}^B > 0\}$ & \multirow{2}{0.01em}{$\biggl\{$} &$ i \in \{x, y, z\},$ \\ 
% 		$a_{\text{min}_{i}}^B = \max\{\beta_l a_{l_{i}}^B \;|\; a_{l_{i}}^B < 0\} $& &$ l \in \{1, \dots, 4\}.$
%     \end{tabular}
% \end{equation}
\changed{\begin{equation}\label{eq:new_acc_bounds}
    \begin{aligned}
        a_{\text{max}}^i = \min\{\beta_l a_{l}^i \;|\; a_{l}^i > 0\}, &\qquad  i \in \{x, y, z\},\\
        a_{\text{min}}^i = \max\{\beta_l a_{l}^i \;|\; a_{l}^i < 0\},  &\qquad l \in \{1, \dots, 4\}.
    \end{aligned}
\end{equation}} 
The described process does not guarantee that $\lVert\mathbf{a}^T\rVert~=~a_{\text{max}}^T $ after recomputing the trajectory with the new per-axis acceleration bounds \refeq{eq:new_acc_bounds}.
Yet as shown in following sections, after few iterations of the algorithm we can approximate the acceleration distribution so that
$\lvert\lVert \mathbf{a}^T \rVert - a_{\text{max}}^T\rvert < \varepsilon_a$ 
holds for a predefined threshold $ \varepsilon_a \in \mathbb{R} $.
This is also the stopping criteria of the algorithm, described in \refalg{alg:thrust_decomposition}.
\vspace{-1em}
\begin{algorithm} 
% \scriptsize
\footnotesize
        \DontPrintSemicolon
	\caption{Limited Thrust Decomposition}\label{alg:thrust_decomposition}
        \SetKwInOut{Input}{In}
        \SetKwInOut{Output}{Out}
	\Input{$ \mathbf{X}_0 $ - trajectory initial conditions, $ \varepsilon_a $ - precision, $ {a}^{T}_{\text{max}} $ - maximal thrust acceleration}
	\Output{${\Pi}$ - resulting trajectory}
	\hrule
 
	$ \changed{\mathbf{a}_{\text{max}}}, \changed{\mathbf{a}_{\text{min}}} \gets \mathbf{getInitAccLimits}({a}^T_{\text{max}}) $ \Comment{\refeq{eq:ltd_acc_limits}}\;
	$ \Pi \gets \mathbf{pmmTraj3D}(\mathbf{X}_0 , \changed{\mathbf{a}_{{\text{max}}}}, \changed{\mathbf{a}_{\text{min}}}) $ \;
	\While{$\lvert\max\{\lVert\mathbf{a_l}^T\rVert\} - a_{\text{max}}^T\rvert > \varepsilon_a$}{
        $ \mathbf{d_1}, \dots, \mathbf{d_4} \gets \mathbf{getDrag}()$ \Comment{\refeq{eq:drag}}\;
		$ \beta_1, \dots, \beta_4 \gets \mathbf{calculateScalingFactors}() $	\Comment{\refeq{eq:ltd_beta2}} \;
		% $\mathbf{a}_{\text{L1}_{j+1}}^B \gets \beta_1 \mathbf{a}_{1_j}^T +\mathbf{d_1} +\mathbf{g} $ \\
  %       $\mathbf{a}_{\text{L2}_{j+1}}^B \gets \beta_2 \mathbf{a}_{2_j}^T +\mathbf{d_2} +\mathbf{g} $ \Comment{\refeq{eq:new_acc_bounds}}\;
        $ \changed{\mathbf{a}_{\text{max}}}, \changed{\mathbf{a}_{\text{min}}} \gets \mathbf{getNewAccLimits}()$\Comment{\refeq{eq:new_acc_bounds}}\;
		Update $ \Pi $ given $\changed{\mathbf{a}_{\text{max}}} $, $\changed{\mathbf{a}_{\text{min}}}$
	}
\end{algorithm}
\vspace{-1em}

% When employed for the velocity constrained bang-singular-bang policy, a reduction in total thrust is assured during the intervals of zero acceleration.
\changed{When employed for the velocity constrained bang-singular-bang policy, no additional steps are needed, as the same acceleration bounds are considered.}
However, an additional challenge involves distributing the maximum velocity norm constraint $v_{max}$ across individual axes.
% To address this, the per-axis velocity constraint $\mathbf{v_{m}}=[v_{m_x}, v_{m_y}, v_{m_z}]^\top$ is adjusted in each iteration according to
To address this, the per-axis velocity constraint \changed{$\mathbf{v_{m}}=[v_{m}^x, v_{m}^y, v_{m}^z]^\top$ }is adjusted in each iteration according to
\begin{equation} \label{eq:VD}
    \mathbf{v_{m}} =  \frac{\mathbf{v}^*}{\lVert \mathbf{v}^* \rVert}\cdot v_{max},
\end{equation}
% where \mbox{$\mathbf{v}^* = [v^*_{x}, v^*_{y}, v^*_{z}]^\top$, $v^*_{i}=\max\limits_{t\in[0, T_{\text{tr}}]} \lvert v_{i}(t) \rvert \, \text{for} \,i\in\{x, y, z\}$}, is a vector of largest velocities attained along each axis.
\changed{where {$\mathbf{v}^* = [v^{x*}, v^{y*}, v^{z*}]^\top$, is a vector of largest velocities attained for each axis, $v^{i*}=\max\limits_{t\in[0, T_{{\text{s}}}]} \lvert v^{i}(t) \rvert$ for $i\in\{x, y, z\}$}.}

\subsection{Velocity Optimization}
\label{sect:vo}

%% Intro
In the previous sections, we assumed that all the states necessary for PMM trajectory planning are known.
However, for a multi-waypoint trajectory, this is generally not the case.
Usually, we have full knowledge about the start and end states; but the via-waypoints are defined only by their positions.
% , i.e. only the path is known.
%% Definition
The problem of finding optimal velocities $ \mathbf{v}_i \in \mathbb{R}^n $ for all via-waypoints of a PMM trajectory that minimize the trajectory duration $ T_{\changed{\Pi}}: \mathbb{R}^n \rightarrow \mathbb{R} $ can be formally written as
\begin{equation}\label{eq:vo_optim_problem_formulation}
	\{\mathbf{v}_1^*, \dots, \mathbf{v}_n^*\} = \arg \min_{\mathbf{v}_1, \dots, \mathbf{v}_n \in \mathbb{R}^n} T_{\changed{\Pi}}(\mathbf{v}_1, \dots, \mathbf{v}_n),
\end{equation}
where $ n $ is the number of via-waypoints of the given path from which the trajectory is computed.
The problem is also subjected to all the trajectory feasibility constraints described in \refsec{sect:pmm} and \refsec{sect:ax_sync}.
%% Approach
To solve~\refeq{eq:vo_optim_problem_formulation}, we propose a new iterative approach based on the gradient method \cite{Boyd2004}, previously used to optimize polynomial trajectories~\cite{Richter20160423}.
% Due to the complexity of the approach, we shall describe the process for the bang-bang policy PMM trajectory and refer to the bang-singular-bang policy where appropriate.
% \begin{figure}[htb]
% 	\centering
% 	\includegraphics[width=0.9\linewidth]{fig/vo-pmm-traj-1ax}
% 	\caption{Velocity profile of an example two-segment one-dimensional trajectory.}
% 	\label{fig:pmm_2seg_1ax}
% \end{figure}
% \begin{figure*}[tb]
%     \vspace{-0.6cm}
% 	\centering
%         \subfloat[Velocity profile]{
% 		\includegraphics[width=0.23\textwidth]{fig/vo-pmm-traj-1ax.pdf}
% 		\label{fig:pmm_2seg_1ax}
% 	}
% 	\subfloat[S segment bounds]{
% 		\includegraphics[width=0.23\textwidth]{fig/vo-sc_bounds.pdf}
% 		\label{fig:pmm_s_bounds}
% 	}
% 	\subfloat[M segment bounds]{
% 		\includegraphics[width=0.23\textwidth]{fig/vo-times_sol.pdf}
% 		\label{fig:pmm_m_bounds_time}
% 	}
%         \subfloat[M segment bounds]{
% 		\includegraphics[width=0.23\textwidth]{fig/vo-times_abs.pdf}
% 		\label{fig:pmm_m_bounds_abs}
% 	}
% 	\caption{Visualization of bounds (red vertical lines) on the updated velocity $v_{12}$ of a two-segment single-axis trajectory with the velocity profile (a), given the scaling factor $\gamma$ for an S-segment (b) and the requirement of non-negative time for the M-segment; displayed are the dependency of the sub-segment time $t_{21}$ on velocity $v_{12}$ (c) and the effect of negative $t_{21}$ on the modified trajectory duration function $T^{abs}_{\changed{\Pi}}$ (d) for both possible solutions $s_1$ and $s_2$.}
% 	\label{fig:time_fcns}
%     \vspace{-0.7cm}
% \end{figure*}
\begin{figure}[tb]
    \changed{
    \vspace{-0.6cm}
	\centering
        \subfloat[Methodology scheme]{
		\includegraphics[width=0.28\textwidth, valign=c]{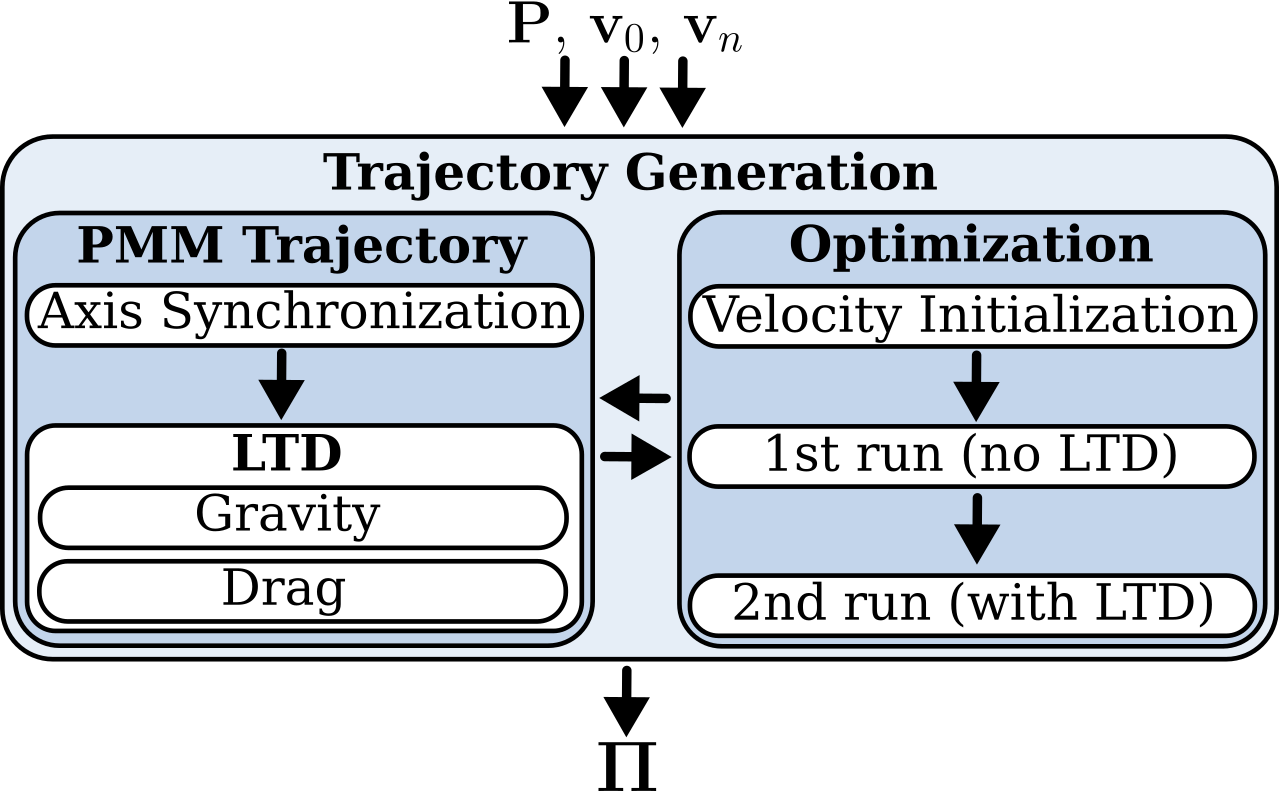}
		\label{fig:method_scheme}
	}
        \subfloat[Velocity profile]{
		\includegraphics[width=0.18\textwidth, valign=c]{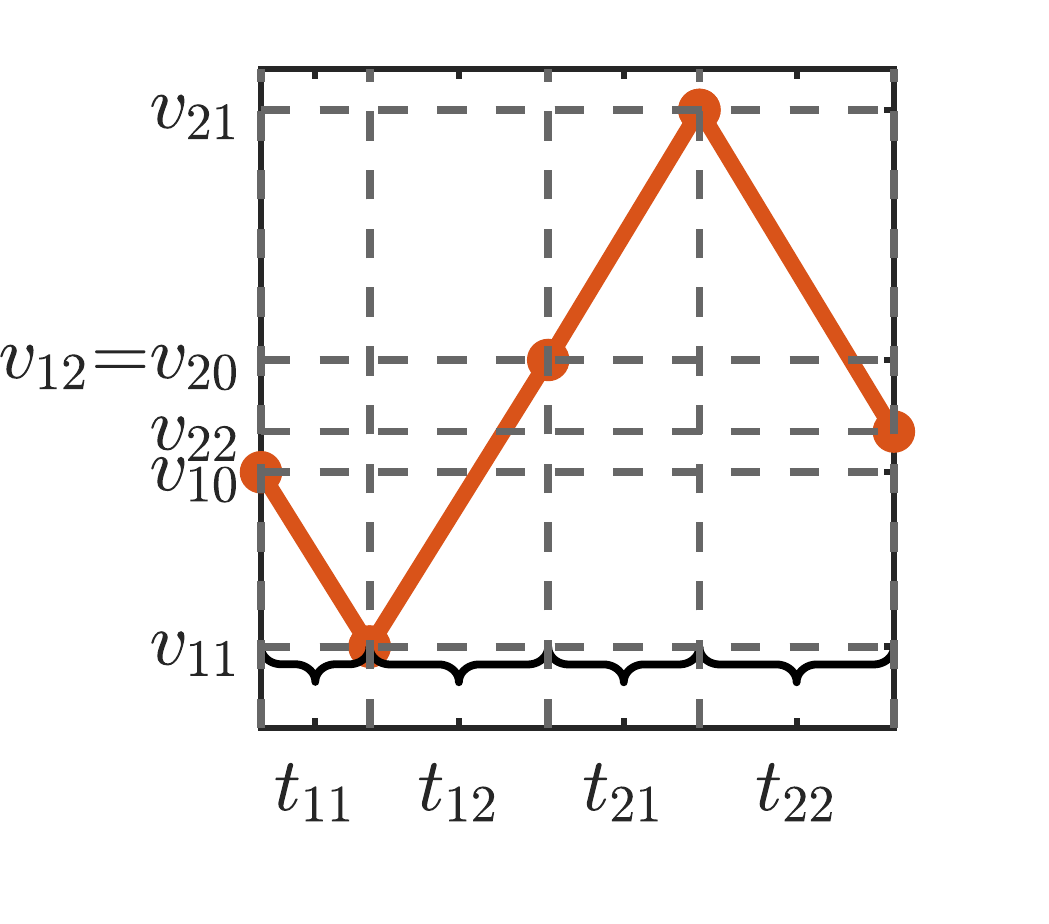}
        \vphantom{\includegraphics[width=0.28\textwidth, valign=c]{fig/method_scheme_v5.png}}
		\label{fig:pmm_2seg_1ax}
	}
	\caption{PMM trajectory generation scheme (a) and the visualization of a two-segment single-axis trajectory velocity profile (b).}
	\label{fig:time_fcns}
 }
    \vspace{-0.7cm}
\end{figure}

%% One-dimensional case
We first demonstrate the core ideas on a two-segment trajectory with defined start and end states and one via-waypoint position.
The velocity profile of such a single-axis trajectory is shown in \reffig{fig:pmm_2seg_1ax}.
The \refsec{sect:pmm} notation is adjusted, where for the second segment $ v_0 = v_{12} $.
The trajectory duration is then equal to
\begin{equation}\label{eq:vo_2seg_tr_duration}
	T_{\changed{\Pi}} = T_1 + T_2 = t_{11} + t_{12} + t_{21} + t_{22},
\end{equation}
\changed{where $T_1$ and $ T_2$ are the corresponding segment durations as defined in (\ref{eq:pmm_tr_time}).}
The problem \refeq{eq:vo_optim_problem_formulation} is then reduced to
\begin{equation}\label{eq:vo_optim_problem_1ax}
	v_{12}^* = \arg \min_{v_{12} \in \mathbb{R}} T_{\changed{\Pi}}(v_{12}) = \arg \min_{v_{12} \in \mathbb{R}} T_1(v_{12}) + T_2(v_{12}).
\end{equation}
The gradient of the objective function $ T_{\changed{\Pi}}(v_{12}) $ is then
\begin{equation}\label{eq:vo_2seg_grad}
	\nabla T_{\changed{\Pi}}(v_{12}) = \dfrac{\partial T_{\changed{\Pi}}}{\partial v_{12}} = \dfrac{\partial T_{1}}{\partial v_{12}} + \dfrac{\partial T_{2}}{\partial v_{12}},
\end{equation}
where the partial derivatives of a segment's duration with respect to its start and end velocity are used.
Having a differentiable objective function and its gradient, we can apply the gradient method to the problem~\refeq{eq:vo_optim_problem_1ax} and formulate the update step for a given step size $ \alpha >0$ as
\begin{equation}\label{eq:vo_optim_1ax_update_step}
	{v}_{{12}_{k+1}} = {v}_{{12}_{k}} - \alpha \, \nabla T_{\changed{\Pi}}({v}_{{12}_k}).
\end{equation}
Velocity optimization is then computed by iteratively optimizing an initial estimate of the unknown velocity $ v_{{12}_0} $ using the update step \refeq{eq:vo_optim_1ax_update_step} until the stopping criterion $|T_{\changed{\Pi}}({v}_{{12}_{k+1}})-T_{\changed{\Pi}}({v}_{{12}_k})| < \varepsilon_T$ is satisfied for a given threshold $\varepsilon_T > 0$.

%% Multi-dimensional case
% To account for the bounded segment duration for all trajectory axes in a multidimensional case, we distinguish between the following roles of single-axis trajectories: a PMM minimum time trajectory segment \refeq{eq:pmm_1seg} will be called an M segment; the remaining synchronization segments will be labeled S segments.
% In the S-S case, meaning both consecutive segments are synchronization trajectories, the boundary velocity $v_{12}$ is not updated as segment durations are determined by other axes and will not be affected by the change.
To account for the bounded segment duration for all axes in a multidimensional case, we distinguish between the following roles of single-axis trajectories: the slowest trajectory \refeq{eq:pmm_1seg} dictating the total time will be called an M segment; the remaining synchronization segments will be labeled S segments.
Since S segments are slowed down, performing optimization of the boundary velocity $v_{12}$ does not yield improvement in the trajectory time when both consecutive segments are synchronization trajectories (S-S case).
The M-S and S-M cases are treated analogously; we showcase only the S-M case.
As the S segment has a constant duration $T_{\text{S}}$, the gradient \refeq{eq:vo_2seg_grad} is reduced to
\begin{equation}\label{eq:vo_sync-min_grad}
	\begin{aligned}
		\nabla T_{\text{S-M}}(v_{12}) &= \dfrac{\partial T_{\text{S-M}}}{\partial v_{12}} = \underbrace{\dfrac{\partial T_{\text{S}}}{\partial v_{12}}}_{0} + \dfrac{\partial T_{\text{M}}}{\partial v_{12}} = \dfrac{\partial T_{\text{M}}}{\partial v_{12}},
	\end{aligned}
\end{equation}
where $T_{\text{M}}$ is the M segment duration.
The S segment places bounds on the final updated velocity as its feasibility must be preserved, which means that the acceleration scale is within the defined interval $\gamma \in (0,1] $.
We do this by fixing $ \gamma = 1 $ and solving the S segment system of equations described in \refsec{sect:ax_sync} for the unknown velocity $v_{2}$, which can be done in closed form.
% There are four possible solutions, from which the ones satisfying the trajectory-feasibility conditions presented in \refsec{sect:ax_sync} and closest to the current boundary velocity $ v_{{12}_k} $ in terms of Euclidean distance are selected as the lower and upper bounds as shown in \reffig{fig:pmm_s_bounds}.
There are four possible solutions. 
% The ones that meet the trajectory feasibility conditions presented in \refsec{sect:ax_sync} and are closest to the current boundary velocity $ v_{{12}_k} $ in Euclidean distance are selected as the lower and upper bounds, as shown in \reffig{fig:pmm_s_bounds}.
\changed{The ones that meet the trajectory feasibility conditions presented in \refsec{sect:ax_sync} and are closest to the current boundary velocity $ v_{{12}_k} $ in Euclidean distance are selected as the lower and upper bounds.}
We refer to S segment velocity bounds computation as $ \mathbf{getSLim} $ in \refalg{alg:velocity_optimization}.
%% MIN bounds
The M segment also places bounds on the velocity update, which are determined by the real-world requirement on the non-negative sub-segment durations.
To find those, we define the following modified trajectory duration function
\begin{equation}\label{eq:vo_tr_duration_abs}
	T_{\changed{\Pi}}^{abs} = |t_1| + |t_2|.
\end{equation}
A real absolute value function $ f(x) = |x| $ is continuous and differentiable everywhere except for $ x = 0 $, $ x \in \mathbb{R} $, which is precisely where $ x $ changes its sign.
By finding the non-differentiable points of the function \refeq{eq:vo_tr_duration_abs}, we can find the values of the boundary velocity, for which the sub-segment durations change their sign and cause trajectory infeasibility.
% This is done by analyzing the domain of partial derivative of \refeq{eq:vo_tr_duration_abs} with respect to the boundary velocity $v_{12}$, resulting in six non-differentiable points.
This is done by finding the six undefined points of the partial derivative of \refeq{eq:vo_tr_duration_abs} with respect to the boundary velocity $v_{12}$.
The relevant ones are selected analogously to the S segment case and returned by the method $ \mathbf{getMLim} $ in \refalg{alg:velocity_optimization}.
% The correspondence of the sub-segment sign change with the non-differentiable points of \refeq{eq:vo_tr_duration_abs} is shown in \reffig{fig:pmm_m_bounds_time} and \reffig{fig:pmm_m_bounds_abs}.
For the M-M case, the velocity update remains according to \refeq{eq:vo_optim_1ax_update_step}, and is subjected to bounds determined by both the M segments as described above.
%% Bang-singular-bang VO
% The described approach can be analogously applied to bang-singular-bang trajectories, with an additional constraint on the velocity $\lvert v^*_{12_i} \rvert\leq\max(\lvert v_{11_i}\rvert, \lvert v_{21_i}\rvert),\, i\in[x, y, z]$ to ensure that the maximum velocity constraint is not exceeded.
The described approach can be analogously applied to bang-singular-bang trajectories, with an additional constraint on the velocity \changed{$\lvert v^{i*}_{12} \rvert\leq\max(\lvert v_{11}^i\rvert, \lvert v_{21}^i\rvert),\, i\in[x, y, z]$ to ensure that the maximum velocity constraint is not exceeded.}

%% Velocity initialization
% Extending the concept of two-segment trajectory optimization to a multi-segment trajectory results in the following iterative method.
\changed{Extending the concept to a multi-segment trajectory results in the following iterative method illustrated in \reffig{fig:method_scheme}.}
First, the unknown velocities are initialized using $\mathbf{initVel}$ method in \refalg{alg:velocity_optimization}.
To obtain the initial velocity heading at each waypoint, we take the previous and next heading vectors $\mathbf{h}_p$ and $\mathbf{h}_n$.
Given the angle $\theta$ between them and the distances to the previous and next waypoints $l_p$ and $l_n$, the $\mathbf{h}_p$ is rotated around the vector $\mathbf{h}_p \times \mathbf{h}_n$ towards $\mathbf{h}_n$ by the angle
\begin{equation}
    % \theta_n = (0.5+r((l_p/(l_p + l_n))-0.5)) \cdot \theta,
    \theta_n = \frac{1-r}{2}\cdot\theta + \frac{r \cdot l_p}{l_p + l_n} \cdot \theta,
\end{equation}
where $r=0.6$ limits the rotation angle interval.
The velocity norm is estimated by assuming zero velocities in the previous and next waypoints and computing the maximal reachable velocity, which is then scaled according to $\theta$ to account for the sharpness of the turn.
%% Resulting VO method
Once the velocities are initialized, at every iteration, starting from one end of the multi-segment trajectory and going toward the other end, we take two neighboring trajectory segments, apply the per-axis velocity update steps to their boundary velocity, and update them accordingly.
The two-segment window then shifts by one segment.
This is repeated until the end of the trajectory is reached, as shown in \refalg{alg:velocity_optimization}.
The order of the velocity updates is changed in every iteration for better convergence.

The per-axis velocity optimization can cause a trajectory segment duration increase due to a role change between the single-axis trajectories caused by the axis synchronization.
Therefore, we check for the duration increase after each update of a two-segment trajectory.
If it occurs, the axes that caused it are determined by tracking the segment role changes ($ \mathbf{getChangedAxes}$ method in \refalg{alg:velocity_optimization}), the corresponding per-axis gradient method update step size is reduced by a reduction factor $ \eta \in (0,1) $ as $\alpha_i = \eta \,\alpha_i$, and the update process repeats.
This is done until either the duration decreases or the update step $ \alpha_{i} $ is reduced below a threshold $ \zeta \in \mathbb{R} $. 
The update for the given $i$-axis is then terminated using $ \alpha_{i} = 0 $.
After each window shift, the update steps for all axes are reinitialized to a given initial value $ \alpha_{\text{init}} > 0 $.

\vspace{-1em}
\begin{algorithm}
% \scriptsize
\footnotesize
    \DontPrintSemicolon
	% \caption{Point-Mass Model Trajectory Generation with Velocity Optimization}
	\caption{PMM Trajectory Velocity Optimization}
    \label{alg:velocity_optimization}
        \SetKwInOut{Input}{In}
        \SetKwInOut{Output}{Out}
	\Input{$ \mathbf{P} $ - set of path waypoints, ($ \mathbf{v}_0 $, $ \mathbf{v}_n$) - start and end velocity}
    \KwData{$ \alpha $ - step size, $ \eta $ - reduction factor, $\zeta$ - reduction threshold; \qquad $ \varepsilon_T $ - time threshold}
	\Output{$\mathbf{\Pi} $ - Optimized trajectory}
	\hrule
	$ \mathbf{V}_{0} \gets  \mathbf{initVel(\mathbf{P})}  $  \;
	$ \mathbf{\Pi}_{0} \gets $ $ \mathbf{pmmTraj3D}\, \forall$ segments using $ \mathbf{V}_{0} $ and $\mathbf{P}$  \;
	\For{ each iteration $k \in \{1,2,\dots\}$}{
		$ \mathbf{V}_{k} \gets \mathbf{V}_{k-1} $ \;

		Switch {$ order \in \{\{1,\dots,n-2\} ,\{n-2,\dots,1\}\} $}

		\For{each segment $ s \in order $}{
			$ \boldsymbol{\alpha} \gets [\alpha, \alpha, \alpha]^\text{T} $ \Comment{Reinitialize step size}\;
			\For{each axis $ i \in \{0,1,2\}$}{
				% $ {p}_{0},{p}_{12} \gets \mathbf{P}[s-1][i],\mathbf{P}[s][i]$\;
				% $ {p}_{22}, {v}_{0} \gets \mathbf{V}_{k-1}[s-1][i],\mathbf{P}[s+1][i] $\;
				% $ {v}_{12},{v}_{22} \gets \mathbf{V}_{k-1}[s][i],\mathbf{V}_{k-1}[s+1][i] $\;
				$ {v}_{12}\gets \mathbf{V}_{k-1}[s][i] $\;
				$ \Pi_{1}^i, \Pi_{2}^i\gets \mathbf{\Pi}_{k}[s-1][i], \mathbf{\Pi}_{k}[s][i] $\;
				\uIf{$ \Pi_{1}^i$ is S and $\Pi_{2}^i$ is S}{
					continue\;
				} \uElseIf{$ \Pi_{1}^i$ is M and $ \Pi_{2}^i$ is M}{
					$ v_{12_{k}} \gets v_{12} - \boldsymbol{\alpha}[i]\nabla T_{\changed{\Pi}}(v_{12}) $\Comment{\refeq{eq:vo_2seg_grad}}\;
                        % $ \mathbf{v}_{12_{B}} \gets \mathbf{getMLim}$ for $\Pi_{1}^i $ and $\Pi_{2}^i$\;
                        $ \mathbf{v}_{12_{B}} \gets \mathbf{getMLim}$($\Pi_{1}^i $), $\mathbf{getMLim}$($\Pi_{2}^i$)\;
                        % Clip $ v_{12_{k}}$ given  $ \mathbf{v}_{12_{B}} $
				} \uElseIf{$ \Pi_{1}^i$ is M and $ \Pi_{2}^i$ is S}{
					% $ \mathbf{v}_{12_{B1}} \gets \mathbf{getSLim}$ for $\Pi_{2}^i$\;
					% $ \mathbf{v}_{12_{B2}} \gets \mathbf{getMLim}$ for $\Pi_{1}^i$\;
					$ v_{12_{k}} \gets v_{12} - \boldsymbol{\alpha}[i] \nabla T_1(v_{12}) $\Comment{\refeq{eq:vo_sync-min_grad}}\;
                        $ \mathbf{v}_{12_{B}} \gets \mathbf{getSLim}$($\Pi_{2}^i$), $\mathbf{getMLim}$($\Pi_{1}^i$)\;
					% Clip $ v_{12_{k}}$ given $  \mathbf{v}_{12_{B1}} $ and $ \mathbf{v}_{12_{B2}} $
				} \ElseIf{$ \Pi_{1}^i$ is S and $ \Pi_{2}^i$ is M}{
					% $ \mathbf{v}_{12_{B1}} \gets \mathbf{getSLim}$ for $\Pi_{1}^i$\;
					% $ \mathbf{v}_{12_{B2}} \gets \mathbf{getMLim}$ for $\Pi_{2}^i$\;
					% $ v_{12_{k}} \gets v_{12} - \boldsymbol{\alpha}[i] \nabla T_2(v_{12}) $\Comment{\refeq{eq:vo_sync-min_grad}}\;
					% Clip $ v_{12_{k}}$ given $  \mathbf{v}_{12_{B1}} $ and $ \mathbf{v}_{12_{B2}} $ 
                        Analogous to the previous case \Comment{\refeq{eq:vo_sync-min_grad}}\;
				}
                Clip $ v_{12_{k}}$ given  $ \mathbf{v}_{12_{B}} $ and 
				$\mathbf{V}_{k}[s][i] \gets v_{12_{k}}$\;
			}
			Update $\Pi_{1}$,$\Pi_{2}$ using $\mathbf{pmmTraj3D}$ for $\mathbf{V}_{k}[s] $\;

			\If{$ (T_{1_{k}} + T_{2_{k}}) > (T_{1_{k-1}} + T_{2_{k-1}})$}{
				$ j \gets \mathbf{getChangedAxes}() $\;
				$ \boldsymbol{\alpha}[j] \gets \eta \boldsymbol{\alpha}[j] \quad \forall j$  \;
				\If{$\exists j$ s.t. $\boldsymbol{\alpha}[j] < \zeta$}{ $\boldsymbol{\alpha}[j] \gets 0$}
				Update $\Pi_{1}$,$\Pi_{2}$ for $\mathbf{V}_{k-1}[s] $\;
				$ s \gets s-1 $\;
			} \Else{
				$\mathbf{\Pi}_{k}[s-1][i],\, \mathbf{\Pi}_{k}[s][i] \gets \Pi_{1},\,\Pi_{2}$\;
			}
		}
		\If{$\changed{\lvert{T_{\Pi_{k}} - T_{\Pi_{k-1}}}\rvert} < \varepsilon_T$}{{break}}	
	}
\end{algorithm}
\vspace{-1em}

%% Integrating the TD into VO
Including the \refalg{alg:thrust_decomposition} in the $\mathbf{pmmTraj3D}$ in \refalg{alg:velocity_optimization} leads to early termination in local minima in some cases, as the assumption of constant per-axis acceleration limits is used in \refalg{alg:velocity_optimization}.
Therefore, we run the optimization in two steps. 
First, the \refalg{alg:velocity_optimization} is run with constant acceleration limits and the resulting trajectory is then recomputed using the \refalg{alg:thrust_decomposition}. 
In the second step, \refalg{alg:velocity_optimization} is run with incorporated \refalg{alg:thrust_decomposition}.
% \changed{A scheme depicting the full pipeline is shown in \reffig{fig:method_scheme}.}

% \begin{itemize}
%     \item Integration of thrust decomposition into vel. optim.
%     % \item Velocity initialization
%     % \item Pseudo-code of the whole algorithm
% \end{itemize}

% \krystof{Main alg., I can write it, but I will need some consultation on how the bang-singular-bang was included. K.}
% \matej{bang-sing-bang is included in place of bang-bang, if max velocity = MAX SCALAR algorithm acts exactly the same, otherwise max velocity is just applied when calculating velocity bounds as bounds have to be less than abs(max velocity). Otherwise, the pseudocode for the algorithm is unaffected by this change IMO.}
% \robert{BTW make the pseudocode rather more pseudo than in the thesis, in the thesis it was fine for the thesis but here it should be smaller and more high level.... If it helps maybe we can have some more smaller algorithms that are called from it such as syncing, thrust decomposition, GD improvement etc.}
% \matej{also IMO, just ignore the bang-sing-bang for pseudocode}

% \subsubsection{Collision-Free Point-Mass Model Trajectory Generation}

%% --------------------------------------------------------------
%% | NMPC                                                       |
%% --------------------------------------------------------------
% \subsection{Collision-free Model Predictive Control}

% \vspace{-1.2cm}

\section{Results\label{sec:results}}
The proposed approach is verified in the following experiments.
We compare the duration of two-waypoint trajectories generated by our LTD approach against the state-of-the-art method with fixed per-axis constraints~\cite{Meyer2023}.
% Effects of LTD coupled with the inclusion of external forces (gravity, drag) on trajectory duration and tracking error are shown through an ablation study.
Effects of LTD coupled with the inclusion of external gravitational and drag forces on trajectory duration and tracking error are shown through an ablation study.
Our gradient-based multi-waypoint trajectory optimization is evaluated against the state-of-the-art sampling-based Cone Refocusing~\cite{Romero2022PMMReplanningMPCC} and time-optimal planner for full quadrotor model~\cite{Foehn2021TimeOptimal} in terms of computational time and trajectory duration.
Finally, we verify the method in real-world flights in an outdoor environment, where we compare the tracking errors when following trajectories generated by our method and trajectories generated for a full quadrotor model~\cite{Foehn2021TimeOptimal}.
% To evaluate our approach, we first systematically compare each individual component to the corresponding state-of-the-art methods.
% We perform an ablation study to show the significance of each component to the entire system.
% Finally, we validate our system's performance in real-world experiments in an outdoor environment.

All methods are implemented in C++ and simulation experiments are run on AMD Ryzen 7 6800HS CPU, with up to 4.7 GHz. 
% Parameters for the velocity optimization algorithm used are $\alpha=10$, $\zeta=0.2$, $\varepsilon=10^{-3}$ for the first run and $\alpha=35$, $\zeta=0.1$, $\varepsilon=10^{-2}$ for the second run.
Parameters $\alpha=10$, $\zeta=0.2$, $\varepsilon_{T}=10^{-3}$ were used for the first step of the \refalg{alg:velocity_optimization}, and values $\alpha=35$, $\zeta=0.1$, $\varepsilon_{T}=10^{-2}$ were adjusted for the second run.
Precision $\varepsilon_{a}=10^{-2}$ was used in \refalg{alg:thrust_decomposition}.
The UAV used in this study, shown in~\reffig{fig:intro}, has a mass of 1.21 kg, maximum collective thrust of 68 N, and estimated drag coefficients \changed{$\delta^x=0.28$, $\delta^y=0.35$, and $\delta^z=0.7$ obtained by analysis of flight data aided with real-time kinematic (RTK) GPS and using various constant speed flights.}
It is equipped with Khadas VIM3 on-board computer and a civilian-grade GNSS capable of withstanding accelerations up to 4g.
% These parameters were considered when determining the maximum values for velocity and acceleration utilized in our experiments.
Considering these factors, we conducted the experiments with maximum total thrust limited to $a^{T}_{\text{max}} = 3.5$g, which is equivalent to 10.5~N per motor.
% \TODO{Maybe we should add the info about MRS system and the NMPC to the place where you describe the UAV (i.e. beginnig of the results?)}
We use the MRS UAV system~\cite{Baca2021mrs,hert2023mrs} for both simulation and real-world validation of the proposed method.
\changed{
The trajectories were tracked using NMPC used in~\cite{Foehn2021TimeOptimal} to track time-optimal trajectories for a full quadrotor model. 
NMPC generates the desired body rates and collective thrust, subsequently sent to the PX4 flight controller.
% The controller was used in~\cite{Foehn2021TimeOptimal} to track time-optimal trajectories for a full quadrotor model.
}

\subsection{Trajectory planning evaluation}
% \begin{itemize}
%     \item Show the scenarios
    % \item Ablation study and influence of different approaches to velocity/acceleration optimization
    % \item study what happens without adding gravity and drag!, i.e. either we need to be conservative or the drone crashes...., also add ablation with without thrust decomposition
    % \item Comparison with the PMM with refocusing
    % \item compare with time-optimal trajectory from foehn, show error when following it
    % \item for our approach show also the time when followed by NMPC and error when following it
    % \item compare single segment with TOP-UAV to show benefit of thrust \& velocity decomposition in contrast to having preset cases equal for both 
% \end{itemize}

\subsubsection{Single-segment trajectory}\label{subsec:single}
% To highlight the impact of our Limited Thrust Decomposition (LTD) approach, we assess the average duration of single-segment trajectories. 
\changed{To assess the impact of our LTD approach, we evaluate the average duration of single-segment trajectories.}
Start and end positions are randomly sampled within a (15~m)$^3$ spatial cuboid, while start and end velocities are sampled within the interval $[-\frac{v_{max}}{\sqrt{3}}, \frac{v_{max}}{\sqrt{3}}]$ for each axis. 
The evaluation involves generating 100 trajectories for each combination of velocity and acceleration limits.
% Furthermore, we discretize the maximum allowed velocity $v_{max}$ and acceleration $a^T_{\text{max}}$ within the range of 0.4 to 40 with a step of 0.4, measured in m/s and m/s$^2$, respectively.

\begin{figure}[!h]
\vspace{-0.6cm}
	\centering
        \subfloat[PMM$_{equal}$]{
		\includegraphics[width=0.45\columnwidth]{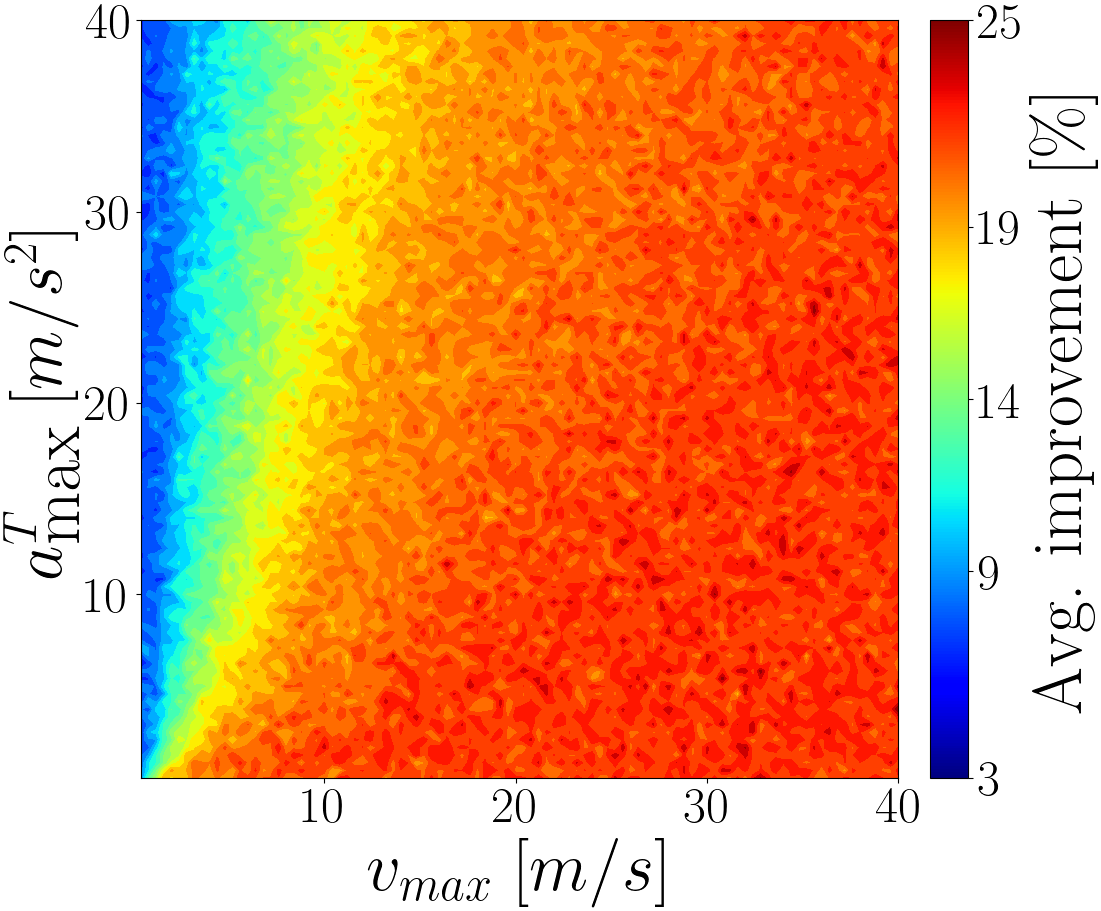}
		\label{fig:vsbasic}
	}
	\subfloat[TOP-UAV$^{\text{++}}$]{
		\includegraphics[width=0.45\columnwidth]{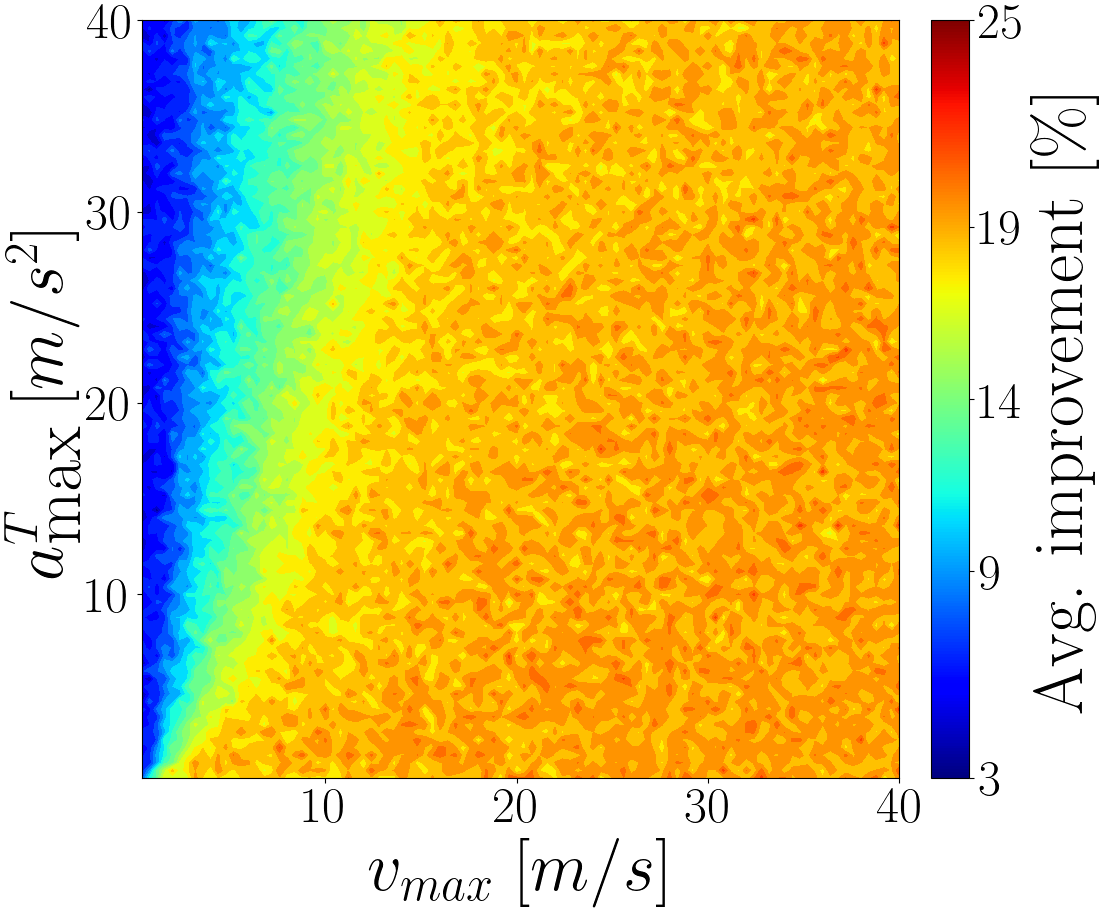}
		\label{fig:vstopauv}
	}
    \vspace{-0.15cm}
	\caption{Improvement of LTD compared to PMM$_{equal}$ and TOP-UAV$^{\text{++}}$.}
	\label{fig:SScomparison}
 \vspace{-0.4cm}
\end{figure}

% Figure~\ref{fig:SScomparison} depicts the average reduction in trajectory duration, expressed as $(T_{\text{other}}-T_{LTD})/T_{\text{other}}$, for LTD compared to PMM$_{equal}$, which is PMM using equal per-axis acceleration and velocity constraints, and TOP-UAV$^{\text{++}}$~\cite{Meyer2023}. 
% For fair comparison, we set $g=0$ and $\mathbf{D}=\mathbf{0}$.
% The average reduction in trajectory duration ranges from 5\% to 25\% when compared to PMM$_{equal}$, and from 3\% to 22\% compared to TOP-UAV$^{\text{++}}$. 
\changed{
Figure~\ref{fig:SScomparison} shows how much LTD reduces trajectory duration compared to PMM using equal per-axis constraints (PMM$_{equal}$) and TOP-UAV$^{\text{++}}$~\cite{Meyer2023}. The reduction is expressed as $(T_{\text{other}}-T_{LTD})/T_{\text{other}}$, where $T_{\text{other}}$ is the duration of the trajectory from the other methods. For a fair comparison, we set $g=0$, $\mathbf{D}=\mathbf{0}$.
Results show that LTD shortens the trajectory duration by 5\% to 25\% compared to PMM$_{equal}$, and by 3\% to 22\% compared to TOP-UAV$^{\text{++}}$.}
% Notably, the least improvement is observed in scenarios with high accelerations and simultaneously low velocities, where TOP-UAV$^{\text{++}}$ managed to generate a faster trajectory than our TD method, in up to 8\% of test cases.
% Notably, the smallest improvement is observed in scenarios with high accelerations and simultaneously low velocities, where in up to 8\% of cases our TD method failed in generating a faster trajectory than TOP-UAV$^{\text{++}}$.
% In 0-8\% of test cases within these scenarios, TOP-UAV$^{\text{++}}$ even generated faster trajectories than our TD method. 
% This situation arises when the initial per-axis maximum velocity $v_{m_i}$ is reached in all axes during the first iteration of the TD algorithm, hindering any change in velocity redistribution as per~\refeq{eq:VD}, in addition to the acceleration-to-velocity ratio being too high for TD to have a significant impact.
% Conversely, our TD approach demonstrates more significant improvement in scenarios characterized by high velocities and accelerations, generating trajectories that are over 18\% faster compared to both PMM$_{equal}$ and TOP-UAV$^{\text{++}}$.
The most significant improvement is observed in scenarios with high velocity and acceleration, aligning with our focus on agile flight.
\changed{In terms of runtimes, the PMM$_{equal}$ approach is the fastest, averaging 464~ns per trajectory.
% This value is associated with generating a single trajectory with a fixed velocity and acceleration distribution.
TOP-UAV$^{\text{++}}$ takes about 980~ns and LTD averages 2.3~$\mu$s per trajectory, which corresponds with average five iterations of LTD to converge.}
Despite a slight increase in runtime, LTD remains suitable for online computation, while significantly reducing trajectory duration, especially in agile flight conditions.
% looks good to me, Great! :)

% \begin{table}[!h]
% \vspace{-0.2cm}
% \centering
% \caption{Average runtime of a single trajectory generation\matej{could be removed actually}}
% \vspace{-1em}
% \begin{tabularx}{\columnwidth}{
%                 >{\centering\arraybackslash}X 
%                 >{\centering\arraybackslash}X  
%                 >{\centering\arraybackslash}X 
%                 >{\centering\arraybackslash}X}  
% \toprule
%   & PMM$_{equal}$ & TOP-UAV$^{\text{++}}$ & \textbf{TD}\\
% \midrule
% runtime [ns] & 464 $\pm$ 125 & 980 $\pm$ 344 & 2338 $\pm$ 997 \\
% \bottomrule 
% \end{tabularx}
% \label{table:runtimes} 
% \vspace{-0.35cm}
% \end{table}

% Given that the runtime of our approach remains within margins suitable for online computation, the minor increase in runtime is outweighed by a significant reduction in trajectory duration, particularly in agile flight conditions.

\begin{figure*}[tb]
    \vspace{-0.6cm}	
    \subfloat[\textit{race}]{
		\includegraphics[width=0.185\textwidth,trim={0.5cm 0 0.5cm 0},clip]{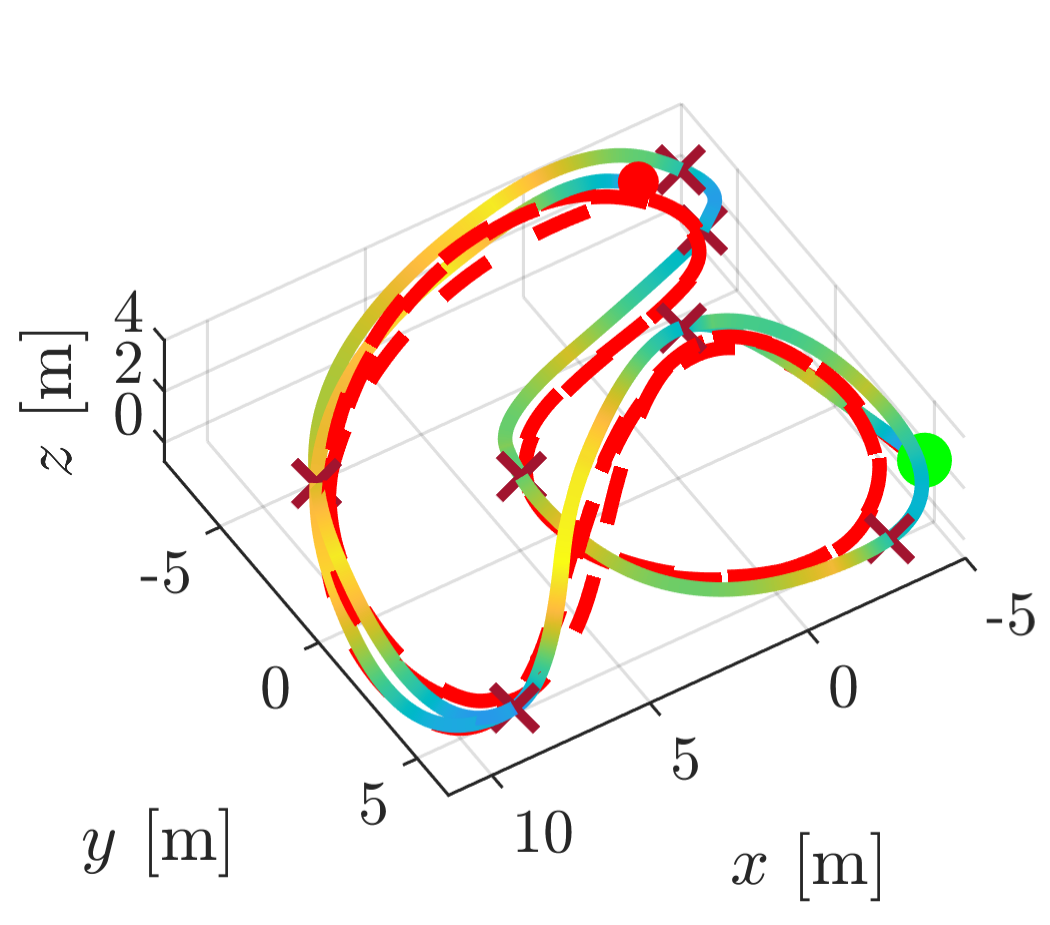}
		\label{fig:race}
	}
	\centering
        \subfloat[\textit{eight}]{
		\includegraphics[width=0.185\textwidth,trim={0.5cm 0 0.5cm 0},clip]{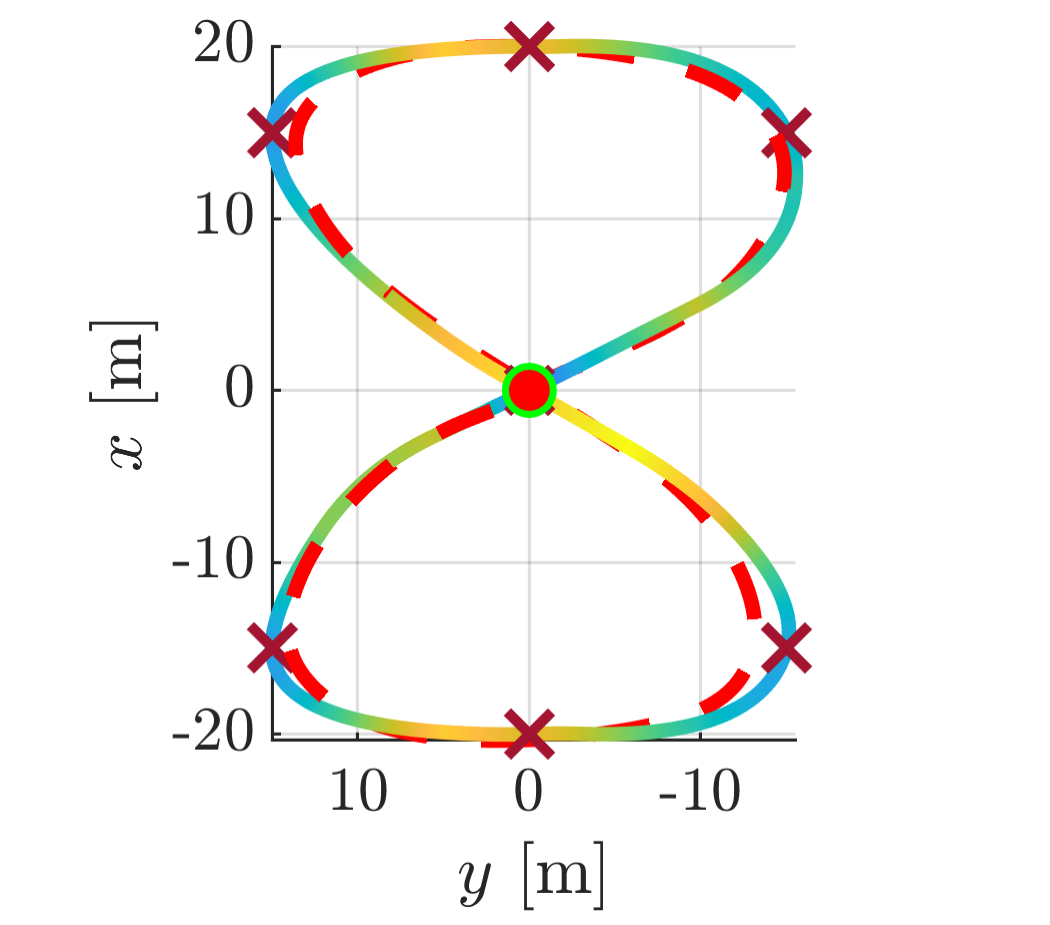}
		\label{fig:eight}
	}
	\subfloat[\textit{cuboid}]{
		\includegraphics[width=0.185\textwidth,trim={0.5cm 0 0.5cm 0},clip]{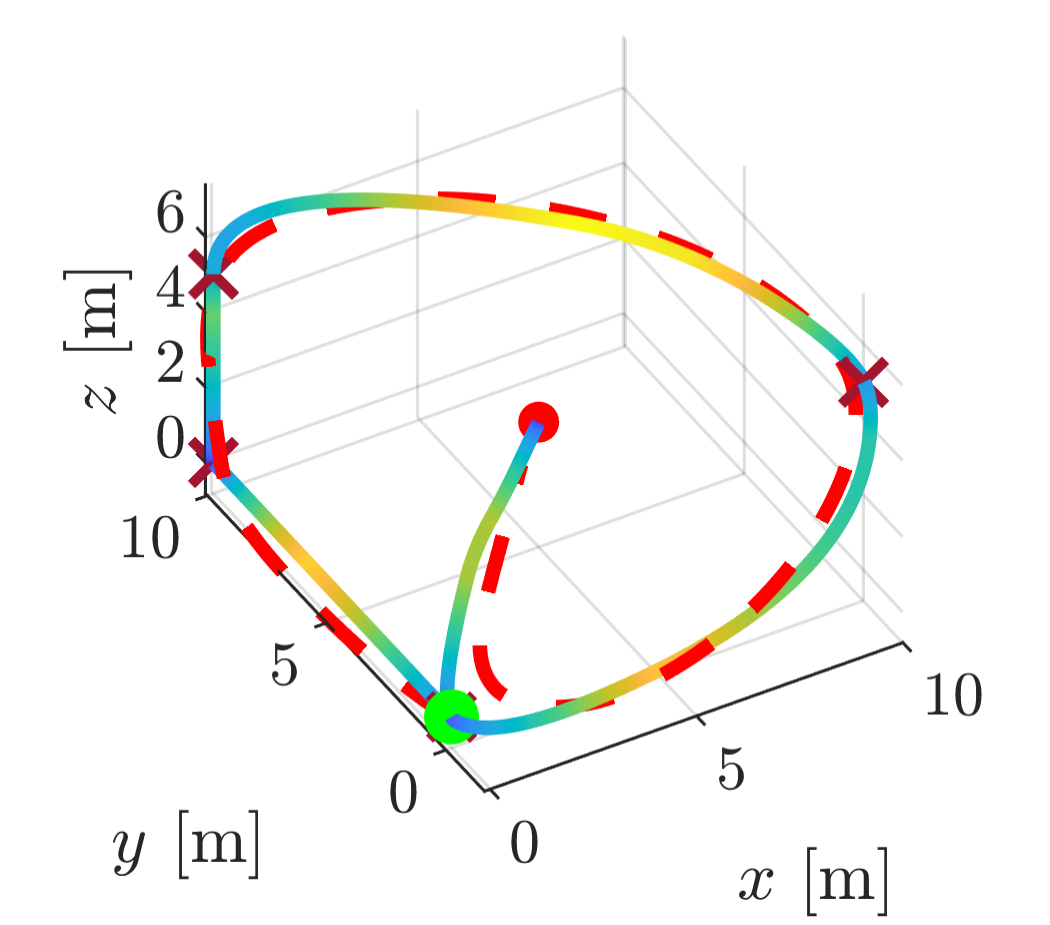}
		\label{fig:cuboid}
	}
	\subfloat[\textit{slalom}]{
		\includegraphics[width=0.185\textwidth,trim={0.5cm 0 0.5cm 0},clip]{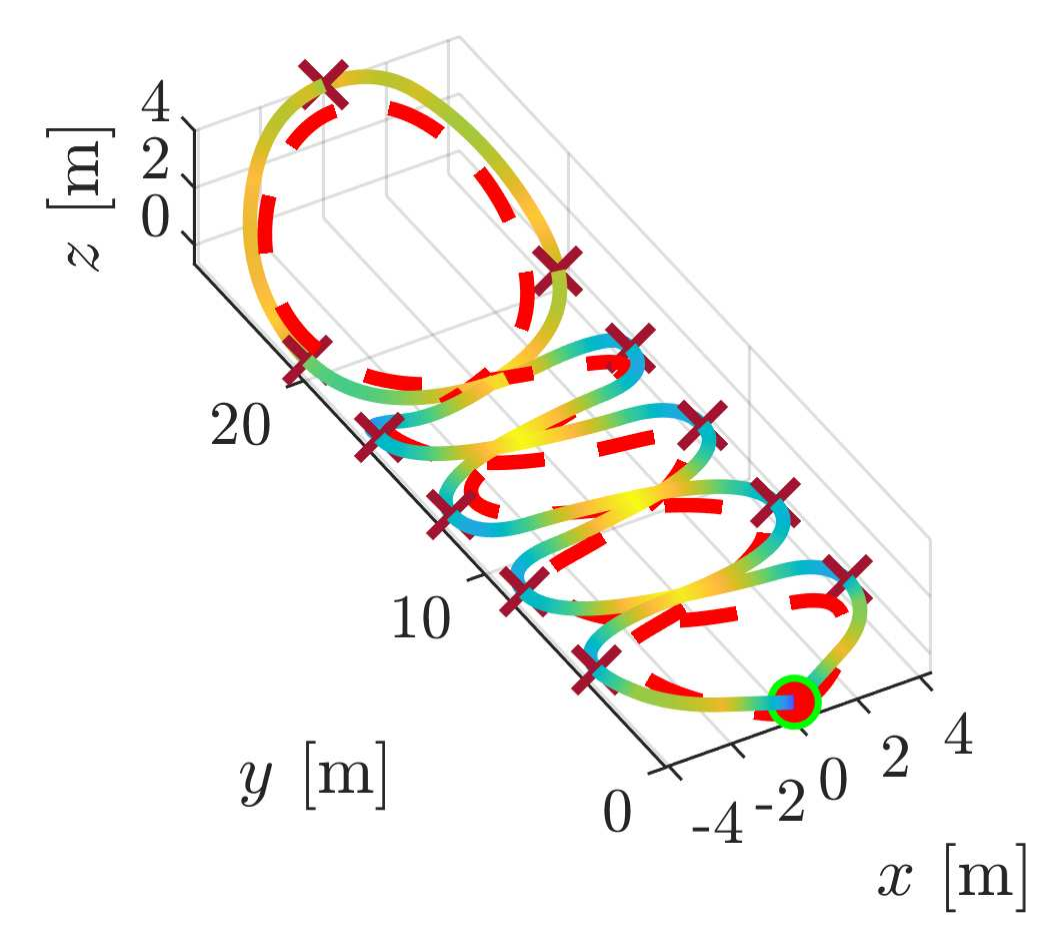}
		\label{fig:slalom}
	}
        \subfloat[\textit{hypotrochoid}]{
		\includegraphics[width=0.185\textwidth,trim={0.5cm 0 0.5cm 0},clip]{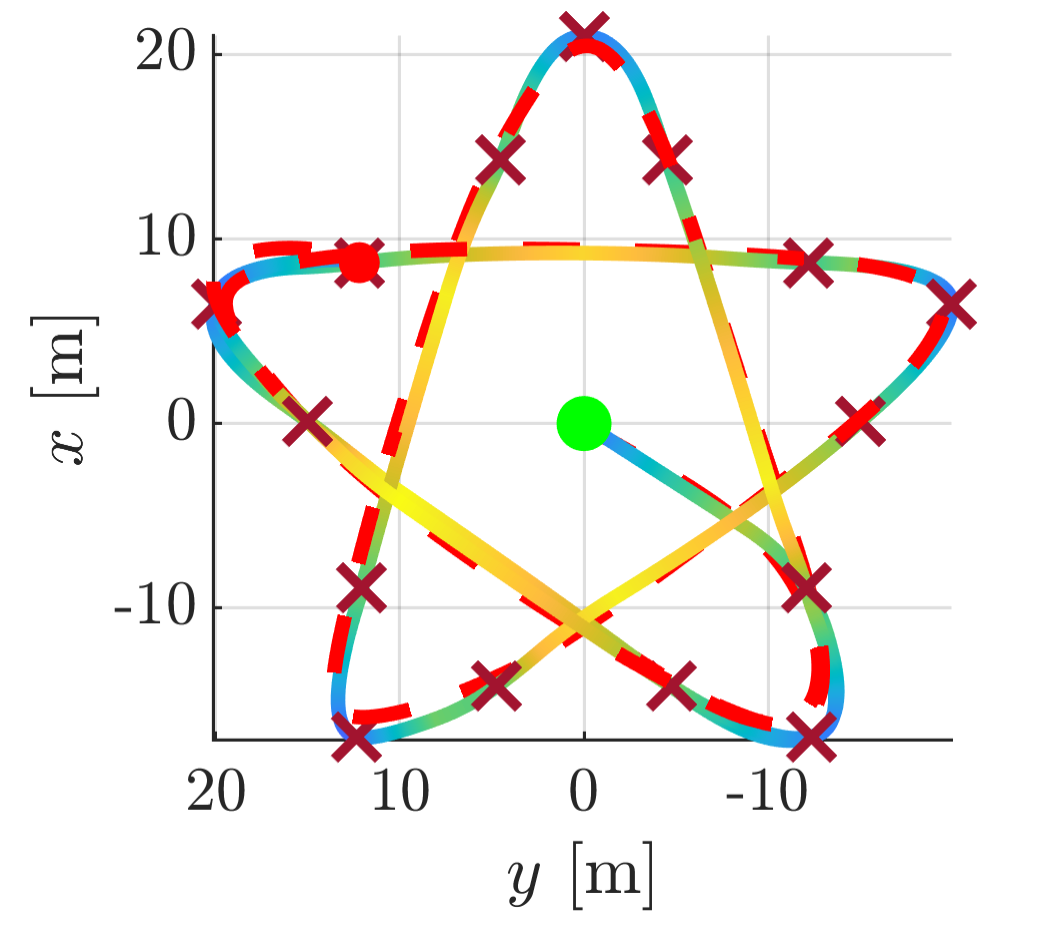}
		\label{fig:hypotrochoid}
	}
 \vspace{-0.15cm}
	\caption{Tested maps defined by their respective start (green point), end (red point), and via (purple crosses) waypoints. The computed trajectory (multi-color line) and the recorded UAV positions from the real-world flight (red dashed line) tracking the trajectories are displayed.}
	\label{fig:map_viz}
    \vspace{-0.7cm}
\end{figure*}

\subsubsection{Multi-waypoint trajectory}

% \begin{itemize}
%     \item Comparison with the PMM with refocusing and time optimal (foehn), ?polynomial?
%     \item comparison when following with NMPC for all
% \end{itemize}
% \TODO{mention acc=3.5g/thurst=10.5N per motor limits, no drag, yes gravity} 
% \matej{I noticed I generated trajs with g=0, so I rerun experiments with gravity and adjusted values in text accordingly as well.}
The proposed Gradient Descent (GD) approach for PMM trajectory optimization in \refalg{alg:velocity_optimization} was compared with the sampling-based cone-refocusing (CR) method~\cite{Romero2022PMMReplanningMPCC} and CPC approach~\cite{Foehn2021TimeOptimal} which considers a full dynamical model of the UAV.
The five maps used for the comparison are shown in~\reffig{fig:map_viz}.
Since CR does not consider drag, we excluded it from this experiment for both our approach and CPC.
The tolerance of missing the waypoint for CPC was set to 0.3~m.
The resulting computational times and trajectory durations are shown in table \reftab{table:VO_CR_comparison}.
% Comparing the trajectory duration, the proposed method achieves better results in 2/5 cases and 3.33\% slower trajectories in average for the rest.
Our method outperforms CR in 2/5 cases in terms of trajectory duration, by 3.15\% on average.
In the remaining cases, it produces trajectories that are, on average, only 2.25\% slower.
% Comparing trajectory durations to CR, our method achieves better results in 2/5 cases and produces trajectories that are, on average, only 2.25\% slower in the remaining cases.
% This is due to the per-axis velocity optimization, where in some cases, the velocity for a given axis is not adjusted due to the trajectory segment being a synchronization trajectory even though it would decrease the trajectory duration in the global scope.
\changed{This is due to the per-axis velocity optimization, where in some cases, the velocity in a waypoint for a given axis is not adjusted due to the adjacent trajectory segments being S-segments, even though an adjustment would decrease the trajectory duration in the global scope.}
Consequently, better results are achieved for the two trajectories, which are planar and the change in position of the waypoints for the main two trajectory axes are of a similar magnitude.
\changed{Without considering drag, the trajectories generated by the CPC algorithm are slower in all tested scenarios. %, as expected, since CPC accounts for the full dynamical model of the UAV.
The full dynamical model contains constraints which PMM violates, leading to CPC's slower trajectories.
However, during flight, NMPC enforces all of CPC's constraints, even when tracking PMM trajectories.
}
\changed{Notably, {PMM$_{\text{GD}}$} planner utilized 99.92\% of available collective thrust on average throughout all 5 trajectories, dropping to 95.82\% when drag is considered.}
% When looking at the computational times of both compared methods, the main contribution of our approach is evident, where compared to the CR method the trajectories are generated 50 to 530 times faster with the largest differences in the case of paths with increasing waypoint count.
% When looking at the computational times of both compared methods, the main contribution of our approach is evident, where compared to the CR method the trajectories are generated 40 to 250 times faster with the largest differences in the case of paths with increasing waypoint count.
Looking at the computational times of all compared methods, the main contribution of our approach is evident.
The CPC algorithm takes hours to converge, whereas our method can generate a trajectory in milliseconds or even tenths of milliseconds for some trajectories, up to two orders of magnitude faster than CR method.
The computational times under 10 ms prove our approach suitable for real-time trajectory planning, essential for real-world deployment in presence of uncertainties, a capability not achievable with current state-of-the-art methods.

\begin{table}[!h]
    \scriptsize
    \vspace{-0.3cm}
    \centering
    \caption{\label{table:VO_CR_comparison} Multi-waypoint trajectory optimization} 
    \vspace{-1em}
    \begin{tabularx}{\columnwidth}{
                    >{\centering\arraybackslash}r|
                    >{\centering\arraybackslash}X 
                    >{\centering\arraybackslash}X|
                    >{\centering\arraybackslash}X 
                    >{\centering\arraybackslash}X|
                    >{\centering\arraybackslash}X 
                    >{\centering\arraybackslash}X}  
    \toprule
      & \multicolumn{2}{c|}{CPC~\cite{Foehn2021TimeOptimal}} & \multicolumn{2}{c|}{PMM$_{\text{CR}}$~\cite{Romero2022PMMReplanningMPCC}} & \multicolumn{2}{c}{\changed{PMM$_{\text{GD}}$}}\\
      map (waypoints) & c.t.[s] & $T_{{\text{s}}}$[s] & c.t.[ms] & $T_{{\text{s}}}$[s] & c.t.[ms] & $T_{{\text{s}}}$[s]\\
        \midrule
        %% 4 digits no gravity
        % race (19) & TBD & 18.5469 & 360 & \textbf{15.8727}  & \textbf{5.98} & 15.9963\\
        % eight (9) & TBD & 10.0359 & 90.7 & 8.9735 & \textbf{1.2} & \textbf{8.7520}\\
        % cuboid (6) & TBD & 6.3614 &  73.3 & \textbf{4.6452} & \textbf{0.311} & 4.9736\\
        % slalom (13) & TBD & 13.4141 &  162 & \textbf{10.7711} & \textbf{0.305} & 11.0028\\
        % hypotrochoid (22) & TBD & 16.5923 &  293 & 15.8772 & \textbf{5.29} & \textbf{15.4610}\\
        
        % %% no gravity
        % race (19) & TBD & 18.55 & 360 & \textbf{15.87}  & \textbf{5.98} & 15.99\\
        % eight (9) & TBD & 10.04 & 90.7 & 8.97 & \textbf{1.2} & \textbf{8.75}\\
        % cuboid (6) & TBD & 6.36 &  73.3 & \textbf{4.64} & \textbf{0.311} & 4.97\\
        % slalom (13) & TBD & 13.41 &  162 & \textbf{10.77} & \textbf{0.305} & 11.00\\
        % hypotrochoid (22) & TBD & 16.59 &  293 & 15.88 & \textbf{5.29} & \textbf{15.46}\\
        
        %% with gravity
        race (19) & 4320 & 17.37 & 376 & \textbf{16.32}  & \textbf{6.21} & 16.48\\
        eight (9) & 1390 & 9.46 & 96 & 9.14 & \textbf{1.28} & \textbf{8.93}\\
        cuboid (6) & 225 & 5.89 &  74.6 & \textbf{4.84} & \textbf{0.78} & 5.10\\
        slalom (13) & 977 & 12.20 &  154 & \textbf{11.05} & \textbf{0.58} & 11.18\\
        hypotrochoid (22) & 9720 & 16.13 &  193 & 16.48 & \textbf{5.26} & \textbf{15.82}\\
        
        %% time in ipopt
        % 4320 & 1390 & 225 & 977 & 9720
        %% total time ?
        % 33560 & 8320 & 533 & 8020 & 104180
        
        \bottomrule 
    \end{tabularx}
    \raggedright
   % \changed{The bold values are the best scoring method in each category.}
\vspace{-0.5cm}
\end{table}

\subsection{Ablation study}

% \subsubsection{External forces ablation}
% \begin{itemize}
%    \item trajectory w/ gravity and drag vs w/o
%    \item controller error when following them - hopefully significantly smaller
% % \end{itemize}
% % \subsubsection{Thrust decomposition ablation}
% % \begin{itemize}
%     % \item Show that the algorithm converges within small number of iterations 
%    \item now that we determined that external forces have a positive outcome for generating more feasible trajectories, show what happens if we don't do thrust decomposition
%    \item first subsubsection could also be moved here maybe?
% \end{itemize}
The significance of using acceleration norm constraint instead of per-axis limits, as well as considering gravity and drag-induced acceleration at high speeds is demonstrated through an ablation study.
The study analyzes their effects on computational time, trajectory duration, and tracking root mean square error (RMSE). 
The results, shown in~\reftab{table:ablation}, consider the methods in the following order of columns: PMM$_{equal}$ and LTD without any external forces, PMM$_{equal}$ and LTD with gravity, and LTD with both gravity and drag-induced acceleration.
As observed in~\refsec{subsec:single}, the LTD approach generates significantly faster trajectories than PMM$_{equal}$, albeit with a higher computational time. 
The fastest trajectories for both PMM$_{equal}$ and LTD are recorded when neither gravity nor drag is considered, as expected since the entire thrust can be utilized.
Yet, it directly causes a significantly higher RMSE. 
Incorporating gravity and drag-induced acceleration results in a reduction of RMSE, bringing LTD with drag nearly on par with PMM$_{equal}$ in terms of RMSE, while still generating significantly faster trajectories.

\begin{table*}[!t]
    \scriptsize
    % \vspace{-0.2cm}
    \centering
    {\renewcommand{\tabcolsep}{4.5pt} 
    \caption{Impact of gravity, LTD and drag on \changed{PMM$_{\text{GD}}$} computational time, trajectory duration and tracking RMSE~[$\textup{m}$]}
    \vspace{-1em}
    \begin{tabularx}{\textwidth}{
                    >{\centering\arraybackslash}r|
                    >{\centering\arraybackslash}X 
                    >{\centering\arraybackslash}X
                    >{\centering\arraybackslash}X|
                    >{\centering\arraybackslash}X 
                    >{\centering\arraybackslash}X
                    >{\centering\arraybackslash}X|
                    >{\centering\arraybackslash}X 
                    >{\centering\arraybackslash}X
                    >{\centering\arraybackslash}X|
                    >{\centering\arraybackslash}X 
                    >{\centering\arraybackslash}X
                    >{\centering\arraybackslash}X|
                    >{\centering\arraybackslash}X 
                    >{\centering\arraybackslash}X
                    >{\centering\arraybackslash}X}  
    \toprule
      & \multicolumn{3}{c|}{PMM$_{equal}$, g=0} & \multicolumn{3}{c|}{\changed{LTD}, g=0} & \multicolumn{3}{c|}{PMM$_{equal}$} & \multicolumn{3}{c|}{\changed{LTD}} & \multicolumn{3}{c}{\changed{LTD} w/ drag}\\
      map & c.t.[ms] & $T_{\changed{\text{s}}}$[s] & RMSE & c.t.[ms] & $T_{\changed{\text{s}}}$[s] & RMSE & c.t.[ms] & $T_{\changed{\text{s}}}$[s] & RMSE & c.t.[ms] & $T_{\changed{\text{s}}}$[s] & RMSE & c.t.[ms] & $T_{\changed{\text{s}}}$[s] & RMSE\\
        \midrule
        race & 0.24 & 19.34 & 0.60 & 6.00 & 16.00 & 0.67 & 0.42 & 21.30 & 0.60 & 6.21 & 16.48 & 0.65 & 11.9 & 18.51 & 0.63\\
        eight & 0.05 & 11.17 & 0.89 & 1.2 & 8.75 & 1.04 & 0.05 & 12.35 & 0.85 & 1.28 & 8.93 & 1.01 & 5.32 & 10.44 & 0.90\\
        cuboid & 0.03 & 6.11 & 0.61 & 0.31 & 4.97 & 0.74 & 0.03 & 6.75 & 0.54 & 0.78 & 5.10 & 0.73 & 1.06 & 5.79 & 0.62\\
        slalom & 0.12 & 13.36 & 0.55 & 0.30 & 11.00 & 0.64 & 0.12 & 14.87 & 0.52 & 0.58 & 11.18 & 0.62 & 2.7 & 12.40 & 0.57\\
        hypotrochoid & 0.35 & 20.10 & 0.79 & 5.25 & 15.46 & 1.15 & 0.32 & 21.86 & 0.77 & 5.26 & 15.82 & 1.04 & 18.6 & 18.51 & 0.85\\
        \bottomrule 
    \end{tabularx}
    \label{table:ablation} 
    }
    % \raggedright
    % \changed{The bold methods are novel approaches presented in this paper.}
    \vspace{-0.7cm}
\end{table*}

%\subsection{Collision-free replanning evaluation}
%\begin{itemize}
%    \item simulation results, 
%\end{itemize}

\subsection{Real-world evaluation}

% \begin{itemize}
%    \item maybe compare with and without ACC decomposition?
%    \item compare tracking quality with and without drag modeled.....
%    \item compare tracking error for PMM trajectory and minimum-time full state trajectory (foehn?)
%    \item consider some online replanning demo?
% \end{itemize}

Finally, we validate our approach in real-world flights by deploying our UAV in an open outdoor environment, utilizing RTK GPS for state estimation.
% We record RMSEs (column RW) for our PMM approach both with modeling drag-induced acceleration (LTD w/  drag) and without it (LTD), as well as for CPC algorithm~\cite{Foehn2021TimeOptimal}, in~\reftab{tab:realworld}.
\changed{We record RMSEs (column RW) for our PMM approach both with modeling drag (PMM$_{\text{GD}}$ w/  drag) and without it (PMM$_{\text{GD}}$), as well as for CPC algorithm~\cite{Foehn2021TimeOptimal}, in~\reftab{tab:realworld}.}
\changed{\change{\reftab{tab:realworld} also includes} computational times for our PMM methods on the embedded computer onboard our UAV, showing that the computation takes around 3.5 times longer, up to 60 ms, which remains suitable for online deployment.}
As anticipated, PMM without drag modeling exhibits the largest tracking error across all trajectories.
However, PMM with drag-induced acceleration outperforms the CPC algorithm in 3 out of 5 cases, with very competitive results in the remaining two. 
\changed{
The most significant advantages of our approach are observed in trajectories covering larger areas, \textit{eight} and \textit{hypotrochoid}.
% during which the highest velocities are reached, resulting in the largest drag.
These  trajectories reach significantly higher velocities, leading to increased drag.
This further highlights our approach in agile flight conditions.
}
\changed{
% To verify our results, we conducted rigorous simulation experiments, performing 5 flights per method per map while excluding external factors such as wind and state estimation errors. 
% The results, recorded in the SIM column of \reftab{tab:realworld}, align with the findings observed in real-world flights.
To verify obtained results, we perform five flights per method per map in simulation to exclude external factors like wind and state estimation errors. 
The results, shown in the SIM column of \reftab{tab:realworld}, align with real-world findings.
% They indicate that trajectory duration has the most significant impact on RMSE, suggesting that the benefits of computing a computationally intensive trajectory for a full model are minimal.
% Instead, our significantly faster trajectory generator that produces trajectories with comparable tracking errors proves to be a more efficient choice.
% They suggest that trajectory duration has the most significant impact on RMSE, showcasing minimal benefit of computing intensive full-model trajectories. 
% Our faster trajectory generator, producing comparable tracking errors, proves to be a more efficient choice.
They indicate that trajectory duration has the most significant impact on RMSE, suggesting minimal benefit in computing a computationally intensive trajectory for a full model. 
Instead, our significantly faster trajectory planner that generates trajectories with comparable tracking errors proves to be a more efficient choice.

}

\begin{table}[!htb] 
   \scriptsize 
   \changed{
   \centering 
   \vspace{-0.3cm}
   {\renewcommand{\tabcolsep}{2.0pt} 
   \caption{Tracking RMSE~[$\textup{m}$] comparison with state-of-the-art\label{tab:realworld}} 
   \vspace{-1em} 
   \begin{tabular}{r|ccc|cccc|cccc} 
    \toprule
    & \multicolumn3{c|}{CPC\cite{Foehn2021TimeOptimal} w/ drag} & \multicolumn4{c|}{{PMM$_{\text{GD}}$}} & \multicolumn4{c}{{PMM$_{\text{GD}}$} w/ drag} \\ 
    map & $T_s$[s] & SIM & RW & c.t.[ms] & $T_s$[s] & SIM & RW & c.t.[ms] & $T_s$[s] & SIM & RW \\ 
       \midrule

race&18.55&\textbf{0.56}&1.05&25&16.48&0.63&1.07&40.7&18.51&0.62&\textbf{1.04}\\ 
eight&10.04&0.92&1.84&5.1&8.93&1.00&2.97&17.4&10.44 &\textbf{0.90}&\textbf{1.48}\\ 
cuboid&6.36&\textbf{0.55}&\textbf{0.93}&3.2&5.10 &0.72&1.04&3.5&5.79 &0.62&1.00\\ 
slalom&13.42&\textbf{0.39}&\textbf{0.88}&2.3&11.18&0.63&1.08&9.0&12.40&0.57&0.92\\ 
hypotrochoid&16.59&0.99&3.62&20.8&15.82&1.03&4.01&61.8&18.51&\textbf{0.87}&\textbf{1.60}\\ 

\bottomrule 
   \end{tabular} 
   }
\vspace{-0.4cm}
}
   %  \raggedright
   % \changed{The bold methods are novel approaches presented in this paper.}
\end{table} 

\section{Conclusions\label{sec:conclusion}}
% In this letter we introduced a novel approach capable of planning minimum-time multi-waypoint trajectories in real-time, with computational time within milliseconds.
% Our Limited Thrust Decomposition algorithm enables the full actuation potential of the drone and improves the duration of single-segment point-mass trajectories by 20\%.
% directly considering gravitational force to fully utilize the collective acceleration produced by the propellers.
% the first work to address drag compensation in the point-mass model, showing its critical importance when flying at high velocities through an ablation study.
% We verify our approach in outdoor real-flights showing that trajectories, with maximum thrust up to 3.5g and velocities reaching over 100~km/h, produced within few milliseconds by our method can be tracked using Nonlinear Model Predictive Controller~(NMPC) with tracking errors similar or smaller than the minimum-time trajectories found within several hours by methods utilizing a full quadrotor model.

We introduced novel approach capable of planning minimum-time multi-waypoint trajectories for agile flight in real time. 
Our gradient-based method optimizes apriori unknown velocities at the waypoints and converges to optimal solutions within milliseconds. 
% We proposed using acceleration norm constraints, rather than per-axis constraints, through the Limited Thrust Decomposition algorithm, which enables to use the full actuation potential of the drone and improves the duration of single-segment point-mass trajectories by 20\%. 
We proposed using acceleration norm constraints instead of per-axis constraints through the Limited Thrust Decomposition algorithm. 
% This enables the full actuation potential of the drone to be utilized and improves the duration of single-segment point-mass trajectories by up to 25\%.
This enables \changed{the drone to utilize its maximum collective thrust} and improves the duration of single-segment point-mass trajectories by up to 25\%.
We directly consider gravitational force to fully utilize the collective acceleration produced by the motors. 
Notably, our work is the first to address drag compensation in the point-mass model, demonstrating the benefits when flying at high velocities, through an ablation study.
% We verified our approach in real flights, showing that trajectories with maximum thrust of up to 3.5g and velocities reaching over 100 km/h, generated within milliseconds, can be tracked using a Nonlinear Model Predictive Controller (NMPC) with tracking errors similar to or smaller than the minimum-time trajectories found after several hours of computation by state-of-the-art methods utilizing a full quadrotor model.
\changed{We verified our approach in real-world flights when using NMPC, where trajectories with accelerations of up to 3.5g and velocities over 100 km/h, generated within milliseconds, produced tracking errors similar to or smaller than the time-optimal trajectories found after several hours of computation by state-of-the-art methods utilizing a full quadrotor model.}
% \change{In future work, we aim to integrate our method with obstacle avoidance, as a part of online re-planning pipeline.}

% \TODO{}
% \lipsum[1-1]

% \krystof{Is it necessary to cite the thesis to prevent auto-plagiarism?}
% \robert{not really, thesis is usually not considered as a publication and a paper from it is fine ;-)}

\bibliographystyle{IEEEtran}
\bibliography{main.bib}

\vfill

\end{document}